\documentclass[journal, final]{IEEEtran} 

\usepackage{graphicx,epsfig,color,bm}
\usepackage{color,cite}
\usepackage{amsmath,epsfig} 
\usepackage{times}
\usepackage{enumerate,type1cm}
\usepackage{amsfonts}
\usepackage{bm}
\usepackage{amssymb}
\usepackage{relsize}
\usepackage{multirow}
\usepackage[english]{babel}
\def\BibTeX{{\rm B\kern-.05em{\sc i\kern-.025em b}\kern-.08em
    T\kern-.1667em\lower.7ex\hbox{E}\kern-.125emX}}

\newtheorem{thm}{Theorem}

\newtheorem{lem}{Lemma}
\newtheorem{prop}{Proposition}

\newcommand{\figwidth}{3.2in} 
\long\def\symbolfootnote[#1]#2{\begingroup%
\def\thefootnote{\fnsymbol{footnote}}\footnote[#1]{#2}\endgroup}

\newcommand{\beq}{\begin{equation}}
\newcommand{\eeq}{\end{equation}}
\newcommand{\beqa}{\begin{eqnarray}}
\newcommand{\eeqa}{\end{eqnarray}}

\newcommand{\pr}{{p}}

\newcommand{\veca}{\mathbf{a}}

\newcommand{\vecn}{\mathbf{n}}

\newcommand{\vect}{\mathbf{t}}
\newcommand{\vecy}{\mathbf{y}}
\newcommand{\vecw}{\mathbf{w}}
\newcommand{\vecx}{\mathbf{x}}

\newcommand{\vecY}{\mathbf{Y}}
\newcommand{\vecX}{\mathbf{X}}

\newcommand{\matW}{\mathbf{W}}
\newcommand{\matT}{\mathbf{T}}
\newcommand{\bmup}{\bm{\Upsilon}}
\newcommand{\bmphi}{\bm{\Phi}}
\newcommand{\bmSigma}{\bm{\Sigma}}

\newcommand{\matA}{\mathbf{A}}
\newcommand{\matB}{\mathbf{B}}

\newcommand{\bmgm}{\bm{\gamma}}
\newcommand{\bmGm}{\Gamma}

\newcommand{\bmtheta}{\bm{\theta}}

\title{Cram\'{e}r-Rao--Type Bounds for Sparse Bayesian Learning}
\author{
\authorblockN{Ranjitha Prasad and Chandra R. Murthy {\it Senior Member, IEEE}}\\
 \thanks{The authors are with the Dept.\ of Electrical Communication
     Eng.\ at IISc, Bangalore,
     India. (e-mails: \{ranjitha.p, cmurthy\}@ece.iisc.ernet.in)}
     \thanks{Copyright (c) 2012 IEEE. Personal use of this material is permitted. However, permission to use this material for any other purposes must be obtained from the IEEE by sending a request to pubs-permissions@ieee.org.}
}
\begin{document}

\maketitle

\begin{abstract}
In this paper, we derive  Hybrid, Bayesian and Marginalized Cram\'{e}r-Rao lower bounds (HCRB, BCRB and MCRB) for the single and multiple measurement vector Sparse Bayesian Learning (SBL) problem of estimating compressible vectors and their prior distribution parameters. We assume the unknown vector to be drawn from a compressible
Student-$t$ prior distribution. We derive CRBs that encompass the deterministic or random nature of the unknown parameters of the prior distribution and the regression noise
variance. We extend the MCRB to the case where the compressible vector is distributed according to a general compressible prior distribution, of which the generalized Pareto
distribution is a special case. We use the derived bounds to uncover the relationship between the compressibility and Mean Square Error (MSE) in the estimates. Further, we
illustrate the tightness and utility of the bounds through simulations, by comparing them with the MSE performance of two popular SBL-based estimators. It is found that the MCRB is
generally the tightest among the bounds derived and that the MSE performance of the Expectation-Maximization (EM) algorithm coincides with the MCRB for the compressible vector.
Through simulations, we demonstrate the dependence of the MSE performance of SBL based estimators on the compressibility of the vector for several values of the number of
observations and at different signal powers.

\end{abstract}
\section{Introduction}

Recent results in the theory of compressed sensing have generated immense interest in sparse vector estimation problems, resulting in a multitude of successful practical signal
recovery algorithms.  In several applications, such as the processing of natural images, audio, and speech, signals are not exactly sparse, but \emph{compressible}, i.e., the magnitudes
of the sorted coefficients of the vector follow a power law decay \cite{needell_cosamp}. In \cite{cevher_learning} and \cite{gribonval_compressible}, the authors show that random vectors
drawn from a special class of probability distribution functions (pdf) known as \emph{compressible priors} result in compressible vectors. 
Assuming that the vector to be estimated (henceforth referred to as the unknown vector) has a
compressible prior distribution enables one to formulate the compressible vector recovery problem in the Bayesian framework, thus allowing the use of Sparse Bayesian Learning (SBL)
techniques \cite{Sparse_RVM}. In his seminal work, Tipping proposed an SBL algorithm for estimating the unknown vector, based on the Expectation Maximization (EM) and McKay updates
\cite{Sparse_RVM}. Since these update rules are known to be slow, fast update techniques are proposed in \cite{Faul_fast}. A duality based algorithm for solving the SBL cost
function is proposed in \cite{wipfRao_ARD}, and $\ell_1 - \ell_2$ based reweighting schemes are explored in \cite{wipf_l1l2}. Such algorithms have been successfully employed for
image/visual tracking \cite{williams_visualsparse}, neuro-imaging \cite{williams_tracksparse, wipf_neuro}, beamforming \cite{wipf_beamforming}, and joint channel estimation and
data detection for OFDM systems \cite{EMSBL_paper}.\\
\indent Many of the aforementioned papers study the complexity, convergence and support recovery properties of  SBL based estimators (e.g., \cite{Faul_fast, wipfRao_ARD}).  In
\cite{gribonval_compressible}, the general conditions required for the so-called instance optimality of such estimators are derived. However, it is not known whether these recovery
algorithms are optimal in terms of the Mean Square Error (MSE) in the estimate or by how much their performance can be improved. In the context of estimating \emph{sparse} signals,
Cram\'{e}r-Rao lower bounds on the MSE performance are derived in \cite{babadi2009CRLB, ben2010CRLB, jutten2011CRLB}. However, to the best of our knowledge, none of the existing
works provide a lower bound on the MSE performance of \emph{compressible} vector estimation. Such bounds are necessary, as they provide absolute yardsticks for comparative analysis
of estimators, and may also be used as a criterion for minimization of MSE in certain problems \cite{CRLB_optimalhelferty}. In this paper, we close this gap in theory by providing
Cram\'{e}r-Rao type lower bounds on the MSE performance of estimators in the SBL framework.

\begin{figure}[t]
\begin{center}
 \scalebox{0.34}{
 \input{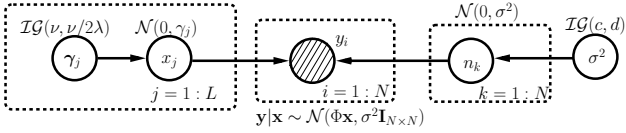}
  }
\caption{Graphical model for SBL: Two stage hierarchical model with the compressible vector taking a conditional Gaussian distribution and the hyperparameters taking an Inverse Gamma distribution. The noise is modeled as white Gaussian distributed, with the noise variance modeled as deterministic/random and  known or unknown.} 
 \label{Graphical_model_system}
 \end{center}
 \end{figure}

As our starting point, we consider a linear Single Measurement Vector (SMV) SBL model given by
\begin{equation}
\vecy = \bmphi\vecx + \vecn,
\label{signal_model}
\end{equation}
where the observations $\vecy \in \mathbb{R}^N$ and the measurement matrix $\bmphi \in \mathbb{R}^{N \times L}$ are known, and $\vecx \in \mathbb{R}^L$ is the unknown sparse/compressible vector to be estimated \cite{SBL_RaoWipf}. Each component of the additive noise $\vecn \in \mathbb{R}^N$ is white Gaussian, distributed as $\mathcal{N}(0,\sigma^2)$, where the variance $\sigma^2$ may be known or unknown. The SMV-SBL system model in \eqref{signal_model} can be generalized to a linear Multiple Measurement Vector (MMV) SBL model given by
\begin{equation}
\matT = \bmphi \mathbf{W} + \mathbf{V}.
\label{mmv_systemmodel}
\end{equation}
Here, $\matT \in \mathbb{R}^{N \times M}$ represents the $M$ observation vectors, the columns of $\matW \in \mathbb{R}^{L \times M}$ are the $M$ sparse/compressible vectors with a common underlying distribution,  and
each column of $\mathbf{V} \in \mathbb{R}^{N \times M}$ is modeled similar to $\vecn$ in \eqref{signal_model} \cite{WipfRao_MMV}. \\ 
\indent In typical compressible vector estimation problems, $\bmphi$ is underdetermined ($N < L$), rendering the problem ill-posed.  Bayesian techniques circumvent this problem by
using a prior distribution on the compressible vector as a regularization, and computing the corresponding posterior estimate. To incorporate a compressible prior in
\eqref{signal_model} and \eqref{mmv_systemmodel}, SBL uses a two-stage hierarchical model on the unknown vector, as shown in Fig.~\ref{Graphical_model_system}. Here, $\vecx \sim
\mathcal{N}(0,\bmup)$, where the diagonal matrix $\bmup$ contains the \emph{hyperparameters} $\bmgm = [\gamma_1, \hdots, \gamma_L]^T$ as its diagonal elements. Further, an Inverse
Gamma (IG) \emph{hyperprior} is assumed for $\bmgm$ itself, because it leads to a Student-$t$ prior on the vector $\vecx$, which is known to be compressible
\cite{Sparse_RVM}.\footnote{The IG hyperprior is conjugate to the Gaussian pdf \cite{Sparse_RVM}.} In scenarios where the noise variance is unknown and random, an IG prior is used
for the distribution of the noise variance as well. For the system model in \eqref{mmv_systemmodel}, every compressible vector $\vecw_i \sim \mathcal{N}(0,\bmup)$, i.e., the $M$
compressible vectors are governed by a common $\bmup$.

It is well known that the Cram\'{e}r-Rao Lower Bound (CRLB) provides a fundamental limit on the  MSE performance of unbiased estimators \cite{Kay} for deterministic parameter
estimation. For the estimation problem in SBL, an analogous bound known as the Bayesian Cram\'{e}r-Rao Bound (BCRB) is used to obtain lower bounds \cite{vantrees_esti_1}, by
incorporating the prior distribution on the unknown vector. If the unknown vector consists of both deterministic and random components, Hybrid Cram\'{e}r-Rao Bounds (HCRB) are
derived \cite{rockah_HCRBfirst}. 

In SBL, the unknown vector estimation problem can also be viewed as a problem involving nuisance  parameters. Since the assumed hyperpriors are conjugate to the Gaussian
likelihood, the marginalized distributions have a closed form and the Marginalized Cram\'{e}r-Rao Bounds (MCRB) \cite{dauwels} can be derived. For example, in the SBL
hyperparameter estimation problem, $\vecx$ itself can be considered a nuisance variable and marginalized from the joint distribution, $\pr_{\vecY,\vecX|\bmgm}(\vecy,\vecx|\bmgm)$,
to obtain the log likelihood as 
\begin{equation}
\log \int_{\vecx}\pr_{\vecY,\vecX|\Gamma}(\vecy,\vecx|\bmgm)\mathrm{d} \vecx = \frac{-(\log |\bm{\Sigma}_y| + \vecy^T \bmSigma_y^{-1} \vecy)}{2},
\label{margi_cost}
\end{equation}
where $\bmSigma_y = \sigma^2 \mathbf{I}_{N \times N}+ \bmphi \bmup \bmphi^T$ \cite{wipfRao_perspective}. 

\begin{figure}[t]
\begin{center}
 \scalebox{0.31}{
 \input{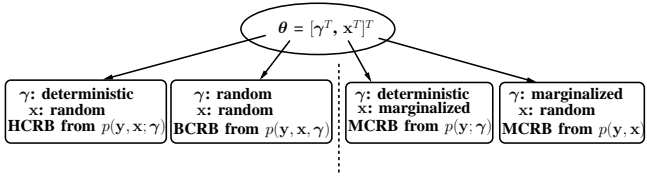}
  }
\caption{Summary of the lower bounds derived in this work when noise variance is assumed to be known.} 
 \label{contributions_noisevarknown}
 \end{center}
 \end{figure}
\begin{figure}[t]
\begin{center}
 \scalebox{0.35}{
 \input{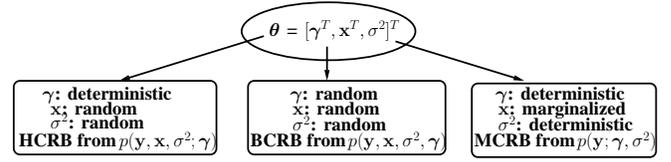}
  }
\caption{Different modeling assumptions and the corresponding bounds derived in this work when noise variance is assumed to be unknown.} 
 \label{contributions_noisevarunknown}
 \end{center}
 \end{figure}

The goal of this paper is to derive Cram\'{e}r-Rao type lower bounds on the MSE performance of estimators based on the SBL framework. Our contributions are as follows:
\begin{itemize}
\item Under the assumption of known noise variance, we derive the HCRB and the BCRB for the unknown vector  $\bmtheta = [\vecx^T, \bmgm^T]^T$, as indicated in the left half of
Fig.~\ref{contributions_noisevarknown}.
\item When the noise variance is known,  we marginalize nuisance variables ($\bmgm$ or $\vecx$) and derive the corresponding MCRB, as indicated in the
right half of Fig.~\ref{contributions_noisevarknown}. Since the MCRB is a function of the parameters of the hyperprior (and hence is an offline bound), it yields insights into the
relationship between the MSE performance of the estimators and the compressibility of $\vecx$. 
\item In the unknown noise variance case,  we derive the BCRB, HCRB and MCRB for the unknown vector $\bmtheta = [\vecx^T, \bmgm^T, \sigma^2]^T$, as indicated in
Fig.~\ref{contributions_noisevarunknown}.
\item We derive the MCRB for a general  parametric form of the compressible prior \cite{gribonval_compressible} and deduce lower bounds for two of the well-known compressible priors, namely, the Student-$t$ and generalized double Pareto distributions.
\item Similar to the SMV-SBL case, we derive the BCRB, HCRB and MCRB for the MMV-SBL model  in~\eqref{mmv_systemmodel}.
\end{itemize}
Through numerical simulations,  we show that the MCRB on the compressible vector $\vecx$ is the tightest lower bound, and that the MSE performance of the EM algorithm achieves this
bound at high SNR and as $N \rightarrow L$. The techniques used to derive the bounds can be extended to handle different compressible prior pdfs used in literature
\cite{cevher_learning}. These
results provide a convenient and easy-to-compute benchmark for comparing the performance of the existing estimators, and in some cases, for establishing their optimality in terms
of the MSE performance.

The rest of this paper is organized as follows.  In Sec.~\ref{prelims}, we provide the basic definitions and describe the problem set up. In Secs.~\ref{LB_sigknown} and
\ref{LB_signotknown}, we derive the lower bounds for the cases shown in Figs.~\ref{contributions_noisevarknown} and~\ref{contributions_noisevarunknown}, respectively. The bounds
are extended to the MMV-SBL signal model in Sec.~\ref{LB_MMVSBL}. The efficacy of the lower bounds is graphically illustrated through simulation results in
Sec.~\ref{sim_res}. We provide some concluding remarks in Sec.~\ref{sec:concl}. In the Appendix, we provide proofs for the Propositions and Theorems stated in the paper.

\textbf{Notation:} In the sequel, boldface small  letters denote vectors and boldface capital letters denote matrices. The symbols $(\cdot)^T$ and $|\cdot|$ denote the transpose
and determinant of a matrix, respectively. The empty set is represented by $\emptyset$, and $\bmGm(\cdot)$ denotes the Gamma function. The function $\pr_{X}(x)$ represents the pdf of
the random variable $X$ evaluated at its realization $x$. Also, $\mbox{diag}(\veca)$ stands for a diagonal matrix with entries on the diagonal given by the vector $\veca$. The
symbol $\nabla_{\bmtheta}$ is the gradient with respect to (w.r.t.) the vector $\bmtheta$. The expectation w.r.t. a random variable $X$ is denoted as $\mathbb{E}_{X}(\cdot)$. Also, 
$\matA \succeq \matB$ denotes that $\matA - \matB$ is positive semidefinite, and $\matA \otimes \matB$ is the Kronecker product of the two matrices $\matA$ and $\matB$.  
\vspace{-1mm}
\section{Preliminaries}
\label{prelims}

As a precursor to the sections that follow,  we define the MSE matrix and the Fisher Information Matrix (FIM) \cite{Kay}, and state the assumptions under which we derive the lower
bounds in this paper. Consider a general estimation problem where the unknown vector $\bmtheta \in \mathbb{R}^n$ can be split into sub-vectors $\bmtheta =
[\bmtheta_r^T,~\bmtheta_d^T]^T$, where $\bmtheta_r \in \mathbb{R}^m$  consists of \emph{random} parameters distributed according to a known pdf, and $\bmtheta_d \in
\mathbb{R}^{n-m}$
consists of \emph{deterministic} parameters. Let $\hat{\bmtheta}(\vecy)$ denote the estimator of $\bmtheta$ as a function of the observations $\vecy$. The MSE matrix
$\mathbf{E}^{\bmtheta}$ is defined as 
\begin{equation}
\mathbf{E}^{\bmtheta} \triangleq \mathbb{E}_{\vecY,\Theta_r}\left[(\bmtheta - \hat{\bmtheta}(\vecy))(\bmtheta - \hat{\bmtheta}(\vecy))^T\right],
\label{error_mat_gen}
\end{equation}
where $\Theta_r$ denotes the random  parameters to be estimated, whose realization is given by $\bmtheta_r$. The first step in obtaining Cram\'{e}r-Rao type lower bounds is to
derive the FIM $\mathbf{I}^{\bmtheta}$ \cite{Kay}. Typically, $\mathbf{I}^{\bmtheta}$ is expressed in terms of the individual blocks of submatrices, where the $(ij)^{\text{th}}$ block is
given by 
\begin{equation}
\mathbf{I}^{\bmtheta}_{ij} \triangleq -\mathbb{E}_{\vecY,\Theta_r}[\nabla_{\bmtheta_i}\nabla_{\bmtheta_j}^T \log \pr_{\vecY,\Theta_r;\Theta_d}(\vecy,\bmtheta_r;\bmtheta_d)].
\label{Infomat_def_gen}
\end{equation}
In this paper, we use the notation  $\mathbf{I}^{\bmtheta}$ to represent the FIM under the different modeling assumptions. For example, when $\bmtheta_r \neq \emptyset$ and
$\bmtheta_d \neq \emptyset$, $\mathbf{I}^{\bmtheta}$ represents a Hybrid Information Matrix (HIM). When $\bmtheta_r \neq \emptyset$ and $\bmtheta_d = \emptyset$,
$\mathbf{I}^{\bmtheta}$ represents a Bayesian Information matrix (BIM). Assuming that the MSE matrix $\mathbf{E}^{\bmtheta}$ exists and the FIM is non-singular, a lower bound on
the MSE matrix $\mathbf{E}^{\bmtheta}$ is given by the inverse of the FIM:
\begin{equation}
\mathbf{E}^{\bmtheta} \succeq \left(\mathbf{I}^{\bmtheta}\right)^{-1}.
\end{equation}
It is easy to verify that the underlying pdfs considered in the SBL model satisfy the regularity conditions required for computing the FIM (see Sec. 5.2.3 in \cite{dauwels}). \\
\indent We conclude this section  by making one useful observation about the FIM in the SBL problem. An assumption in the SMV-SBL framework is that $\vecx$ and $\vecn$ are
independent of each other (for the MMV-SBL model, $\matT$ and $\matW$ are independent). This assumption is reflected in the graphical model in Fig.~\ref{Graphical_model_system},
where the compressible vector $\vecx$ (and its attribute $\bmgm$) and the noise component $\vecn$ (and its attribute $\sigma^2$) are on unconnected branches. Due to this, a
submatrix of the FIM is of the form
\begin{eqnarray}
&\mathbf{I}^{\bmtheta}_{\bmgm \xi} = -\mathbb{E}_{\vecX,\vecY,\Gamma,\Xi}\left[\nabla_{\bmgm}\nabla_{\xi}\left\{\log \pr_{\vecY|\vecX,\Xi}(\vecy|\vecx,\xi)\right.\right.\nonumber\\
&\left.\left.+ \log\pr_{\vecX, \Gamma}(\vecx , \bmgm)+\log\pr_{\Xi}(\xi)\right\}\right],
\end{eqnarray}
where there are no terms in which  both $\bmgm$ and $\xi = \sigma^{2}$ are jointly present. Hence, the corresponding terms in the above mentioned submatrix are always zero. This is
formally stated in the following Lemma.

\begin{lem}
When $\bmtheta_i = \bmgm$ and  $\bmtheta_j = \sigma^2$, the $(ij)^{\text{th}}$ block matrix of the FIM $\mathbf{I}^{\bmtheta}$ given by \eqref{Infomat_def_gen} simplifies
to $\mathbf{I}^{\bmtheta}_{ij} = \mathbf{0}_{L \times 1}$, i.e., to an all zero vector.
\label{prop_gen_noise}
\end{lem}

\section{SMV-SBL: Lower Bounds when $\sigma^2$ is Known}
\label{LB_sigknown}
In this section, we derive lower  bounds for the system model in \eqref{signal_model} for the scenarios in Fig.~\ref{contributions_noisevarknown}, where the unknown vector is
$\bmtheta = [\vecx^T, \bmgm^T]^T$. We examine different modeling assumptions on $\bmgm$ and derive the corresponding lower bounds.
\vspace{-2mm}
\subsection{Bounds from the Joint pdf}

\subsubsection{HCRB for $\bmtheta = [\vecx^T, \bmgm^T]^T$}

In this subsection, we consider  the unknown variables as a hybrid of a deterministic vector $\bmgm$ and a random vector $\vecx$ distributed according to a Gaussian distribution
parameterized by $\bmgm$.  Using the assumptions and notation in the previous section, we obtain the following proposition.

\begin{prop}
\label{prop1}
For the signal model in \eqref{signal_model},  the HCRB on the MSE matrix $\mathbf{E}^{\bmtheta}$ of the unknown vector $\bmtheta = [\vecx^T, \bmgm^T]^T$ with the parameterized
distribution of the compressible signal $\vecx$ given by $\mathcal{N}(0,\bmup)$, and with $\bmgm$ modeled as unknown and deterministic, is given by 
$\mathbf{E}^{\bmtheta} \succeq (\mathbf{H}^{\bmtheta})^{-1}$, where
\begin{eqnarray}
&\mathbf{H}^{\bmtheta} \triangleq
\begin{bmatrix}
\mathbf{H}^{\bmtheta}(\vecx) &   \mathbf{H}^{\bmtheta}(\vecx,\bmgm)\\
(\mathbf{H}^{\bmtheta}(\vecx,\bmgm))^T & \mathbf{H}^{\bmtheta}(\bmgm)
\end{bmatrix} =  \nonumber\\
&\begin{bmatrix}
\left(\frac{\bmphi^T \bmphi}{\sigma^2}  + \bmup^{-1}\right) &   \mathbf{0}_{L \times L}\\
\mathbf{0}_{L \times L} & \textnormal{diag}({2 \gamma_1^2},~{2\gamma_2^2},~\hdots ,~{2\gamma_L^2})^{-1}
\end{bmatrix}.
\label{case1_HCRB}
\end{eqnarray}
\end{prop}
\emph{Proof:}
See Appendix \ref{pf_prop1}.

Note that the lower bound  on the estimate of $\vecx$ depends on the prior information through the diagonal matrix $\bmup$. In the SBL problem, the realization
of the random parameter $\bmgm$ has to be used to compute the bound above, and hence, it is referred to as an online bound. Also, the lower bound on the MSE matrix of $\vecx$
is $\mathbf{E}^{\bmtheta} \succeq \left(\frac{\bmphi^T \bmphi}{\sigma^2}  + \bmup^{-1}\right)^{-1}$, which is the same as the lower bound on the error covariance of the Baye's vector
estimator for a linear model (see Theorems 10.2 and 10.3 in \cite{Kay}), and is achievable by the MMSE estimator when $\bmup = \mbox{diag}(\gamma_1, \hdots, \gamma_L)$ is known. 

\subsubsection{BCRB for $\bmtheta = [\vecx^T, \bmgm^T]^T$}

For deriving the BCRB, a hyperprior distribution is considered on $\bmgm$, and the resulting $\vecx$ is  viewed as being drawn from a compressible prior distribution. The most commonly used
hyperprior distribution in the literature is the IG distribution \cite{Sparse_RVM}, where $\gamma_i, i = 1, 2, \ldots, L$ are distributed as $\mathcal{IG}\left(\frac{\nu}{2}, \frac{\nu}{2\lambda}\right)$,
given by
\begin{equation}
\pr_{\Gamma}(\gamma_i) \triangleq \left(\Gamma\left(\frac{\nu}{2}\right)\right)^{-1} \left(\frac{\nu}{2\lambda}\right)^{\frac{\nu}{2}}{\gamma_i^{\left(-\frac{\nu}{2} -
1\right)}\exp\left\{-\frac{\nu}{2\lambda\gamma_i}\right\}},
\label{IG_gamma}
\end{equation}
where $\gamma_i \in (0,\infty),~~ \nu,\lambda >0$. Using the definitions and notation in the previous section, we state the following proposition.
\begin{prop}
\label{prop2}
For the signal model in \eqref{signal_model}, the BCRB  on the MSE matrix $\mathbf{E}^{\bmtheta}$ of the unknown random vector $\bmtheta = [\vecx^T, \bmgm^T]^T$, where the
conditional distribution of the compressible signal $\vecx|\bmgm$ is $\mathcal{N}(0,\bmup)$, and the hyperprior distribution on $\bmgm$ is $\prod_{i =
1}^{L}\mathcal{IG}\left(\frac{\nu}{2}, \frac{\nu}{2\lambda}\right)$, is given by $\mathbf{E}^{\bmtheta} \succeq (\mathbf{B}^{\bmtheta})^{-1}$, where
\begin{eqnarray}
&\mathbf{B}^{\bmtheta} \triangleq
\begin{bmatrix}
\mathbf{B}^{\bmtheta}(\vecx) &   \mathbf{B}^{\bmtheta}(\vecx,\bmgm)\\
(\mathbf{B}^{\bmtheta}(\vecx,\bmgm))^T & \mathbf{B}^{\bmtheta}(\bmgm)
\end{bmatrix} = \nonumber\\
&\begin{bmatrix}
 \left(\frac{\bmphi^T \bmphi}{\sigma^2}  + {\lambda}\mathbf{I}_{L \times L}\right) &  \mathbf{0}_{L \times L}\\
\mathbf{0}_{L \times L} &  \frac{\lambda^2(\nu + 2)(\nu + 7)}{2\nu}\mathbf{I}_{L \times L}
\end{bmatrix}.
\label{bcrb_gamma}
\end{eqnarray}
\end{prop}
\emph{Proof:}
See Appendix \ref{prop2proof}.

It can be seen from $\mathbf{B}^{\bmtheta}$ that the lower  bound on the MSE of $\hat{\bmgm}(\vecy)$ is a function of the parameters of the IG prior on $\bmgm$, i.e., a
function of $\nu$ and $\lambda$, and it can be computed without the knowledge of realization of $\bmgm$. Thus, it is an offline bound. 
\subsection{Bounds from Marginalized Distributions}
\label{sec_case3}
\subsubsection{MCRB for $\bmtheta = [\bmgm]$}

Here, we derive the MCRB for $\bmtheta = [\bmgm]$, where  $\bmgm$ is an unknown deterministic parameter. This requires the marginalized distribution
$\pr_{\vecY;\bmgm}(\vecy;\bmgm)$, which is obtained by considering $\vecx$ as a nuisance variable and marginalizing it out of the joint distribution
$\pr_{\vecX, \vecY;\bmgm}(\vecx,\vecy;\bmgm)$, to obtain \eqref{margi_cost}. Since $\bmgm$ is a deterministic parameter, the pdf $\pr_{\vecY; \bmgm}(\vecy;\bmgm)$ must satisfy the
regularity condition in \cite{Kay}. We have the following theorem.
\begin{thm}
\label{thm1}
For the signal model in \eqref{signal_model}, the log likelihood  function $\log\pr_{\vecY; \bmgm}(\vecy;\bmgm)$ satisfies the regularity conditions in~\cite{Kay}. Further, the MCRB on the MSE matrix $\mathbf{E}^{\bmgm}$ of the unknown deterministic vector $\bmtheta = [\bmgm]$ is given by $\mathbf{E}^{\bmgm}\succeq({\mathbf{M}^{\bmgm}})^{-1}$, where the
$(ij)^{\text{th}}$ element of $\mathbf{M}^{\bmgm}$ is given by
\begin{equation}
\mathbf{M}^{\bmgm}_{ij} = \frac{1}{2}(\Phi_j^T \bmSigma_y^{-1}\Phi_i)^2,
\label{crlb_case3}
\end{equation}
for $1 \leq i,j \leq L$, where $\Phi_i$ is the $i^{\text{th}}$ column of $\bmphi$, and $\bmSigma_y = \sigma^2 \mathbf{I}_{N \times N}+ \bmphi \bmup \bmphi^T$, as defined earlier.
\end{thm}
\emph{Proof:}
See Appendix \ref{pf_thm1}.

To intuitively understand \eqref{crlb_case3}, we consider a  special case of $\bmphi^T\bmphi = N\mathbf{I}_{N \times N}$, and use the Woodbury formula to simplify
$\bmSigma_y^{-1}$, to obtain the $(ii)^{\text{th}}$ entry of the matrix $\mathbf{M}^{\bmgm}$ as 
\begin{eqnarray}
\mathbf{M}^{\bmgm}_{ii} = 2\left(\frac{\sigma^2}{N} + \gamma_i\right)^{-2}.
\label{marginal_bound_case3}
\end{eqnarray}
Hence, the error in $\gamma_i$ is bounded as  $\mathbf{E}^{\bmgm}_{ii} \geq 2\left(\frac{\sigma^2}{N}+\gamma_i\right)^2$. As $N \rightarrow \infty$, the bound reduces to $2
\gamma_i^2$, which is the same as the lower bound on the estimate of $\bmgm$ obtained as the lower-right submatrix in \eqref{case1_HCRB}. For finite $N$, the MCRB is tighter than the
HCRB. 


\subsubsection{MCRB for $\bmtheta = [\vecx]$}

In  this subsection, we assume a hyperprior on $\bmgm$, which leads to a joint distribution of $\vecx$ and $\bmgm$, from which $\bmgm$ can be marginalized. Further, assuming
specific forms for the hyperprior distribution can lead to a compressible prior on $\vecx$. For example, assuming an IG hyperprior on $\bmgm$ leads to an $\vecx$ with a Student-$t$
distribution. Sampling from a Student-$t$ distribution with parameters $\nu$ and $\lambda$ results in a $\nu$-compressible $\vecx$~\cite{cevher_learning}.
The Student-$t$ prior is given by
\begin{equation}
\pr_{\vecX}(\vecx) \triangleq \left(\frac{\bmGm((\nu+1)/2)}{\Gamma(\nu/2)}\right)^L\left(\frac{\lambda}{\pi\nu}\right)^{\frac{L}{2}}\prod_{i=1}^{L}\left(1 + \frac{\lambda x_i^2}{\nu}
\right)^{-\frac{\nu+1}{2}},
\label{studentt}
\end{equation}
where $x_i \in (-\infty,\infty),~~ \nu,\lambda >0$, $\nu$ represents the number of degrees of freedom and $\lambda$ represents the inverse variance of the distribution. Using the
notation developed so far, we state the
following theorem.
\begin{thm}
\label{thm2}
For the  signal model in \eqref{signal_model}, the MCRB on the MSE matrix $\mathbf{E}^{\vecx}$ of the unknown compressible random vector $\bmtheta = [\vecx]$ distributed as
\eqref{studentt}, is given by $\mathbf{E}^{\vecx} \succeq ({\mathbf{M}^{\vecx}})^{-1}$, where
\begin{equation}
\mathbf{M}^{\vecx} = \frac{\bmphi^T\bmphi}{\sigma^2} + \frac{\lambda(\nu+1)}{(\nu+3)}\mathbf{I}_{L \times L}.
\label{mcrb_case4}
\end{equation}
\end{thm}
\emph{Proof:}
See Appendix \ref{pf_thm2}.

We see  that the bound derived depends on the parameters of the Student-$t$ pdf. From \cite{gribonval_compressible}, the prior is \emph{``somewhat''} compressible for $2<\nu<4$,
and \eqref{mcrb_case4} is nonnegative and bounded for $2<\nu<4$, i.e., the bound is meaningful in the range of $\nu$ used in practice. Note that, by choosing $\lambda$ to be large
(or the variance of $\vecx$ to be small), the bound is dominated by the prior information, rather than the information from the observations, as expected in Bayesian bounds~\cite{Kay}. 

It is conjectured in \cite{dauwels} that, in general, the MCRB is tighter than the BCRB. Analytically comparing the MCRB \eqref{mcrb_case4} with the  BCRB \eqref{case1_HCRB}, we see that for the SBL problem of estimating a compressible vector, the MCRB is indeed tighter than the BCRB, since 
$$\left(\frac{\bmphi^T\bmphi}{\sigma^2} + \frac{\lambda(\nu+1)}{(\nu+3)}\mathbf{I}_{L \times L}\right)^{-1} \succeq \left(\frac{\bmphi^T\bmphi}{\sigma^2} + \lambda \mathbf{I}_{L
\times L}\right)^{-1}.$$ 

The  techniques used to derive the bounds in this subsection can be applied to any family of compressible distributions. In \cite{gribonval_compressible}, the authors propose a
parametric form of the Generalized Compressible Prior (GCP) and prove that such a prior is compressible for certain values of $\nu$. In the following subsection, we derive the MCRB
for the GCP.

\subsection{General Marginalized Bounds}
In this  subsection, we derive MCRBs for the parametric form of the GCP. The GCP encompasses the double Pareto shrinkage type prior \cite{Double_Pareto_artin} and the Student-$t$
prior \eqref{studentt} as its special cases. We consider the GCP on $\vecx$ as follows
\begin{equation}
\pr_{\vecX}(\vecx) \triangleq K^L \prod_{i = 1}^L\left( 1+\frac{\lambda\left|x_i\right|^{\tau}}{\nu}  \right)^{-(\nu+1)/\tau},
\label{gen_compprior}
\end{equation}
where $x_i \in (-\infty,\infty), \tau,\nu,\lambda >0$, and the normalizing constant $K \triangleq
\frac{\tau}{2}\left(\frac{\lambda}{\nu}\right)^{1/\tau}\frac{\Gamma((\nu+1)/\tau)}{\Gamma(1/\tau)\Gamma(\nu/\tau)}$. When $\tau =
2$, \eqref{gen_compprior} reduces to the Student-$t$ prior in \eqref{studentt}, and when $\tau = 1$, it reduces to a generalized double Pareto shrinkage prior
\cite{Double_Pareto_artin, balakrishnan_priors}. Also, the expression for the GCP in \cite{gribonval_compressible} can be obtained from \eqref{gen_compprior} by setting $\lambda = 1$, and defining
$\nu \triangleq s-1$. 
The following theorem provides the MCRB for the GCP.
\begin{thm}
\label{thm3}
For the  signal model in \eqref{signal_model}, the MCRB on the MSE matrix $\mathbf{E}_{\tau}^{\bmtheta}$ of the unknown random vector $\bmtheta = [\vecx]$, where $\vecx$ is
distributed as the GCP in \eqref{gen_compprior}, is given by $\mathbf{E}_{\tau}^{\bmtheta} \succeq ({\mathbf{M}_{\tau}^{\bmtheta}})^{-1}$, where 
\begin{equation}
{\mathbf{M}_{\tau}^{\bmtheta}} = \frac{\bmphi^T\bmphi}{\sigma^2} + T_{\tau},
\label{mcrb_gen}
\end{equation}
where $T_{\tau} = \frac{\tau^2(\nu+1)}{(\nu + \tau + 1)} \left(\frac{\lambda}{\nu}\right)^{2/\tau}\frac{\Gamma\left(\frac{\nu+2}{\tau}\right)\Gamma\left(2-\frac{1}{\tau}\right)}{\Gamma\left(\frac{1}{\tau}\right)\Gamma\left(\frac{v}{\tau}\right)}\mathbf{I}_{L \times L}.$
\end{thm}
\emph{Proof:}
See Appendix \ref{pf_thm4}.\\
\indent It  is straightforward to verify that for $\tau = 2$, \eqref{mcrb_gen} reduces to the MCRB derived in \eqref{mcrb_case4} for the Student-$t$ distribution. For $\tau = 1$,
the inverse of the MCRB can be reduced to
\begin{equation}
{\mathbf{M}_{\tau}^{\bmtheta}} = \frac{\bmphi^T\bmphi}{\sigma^2} + \frac{\lambda^2(\nu+1)^2}{\nu(\nu+2)}\mathbf{I}_{L \times L}.
\label{mcrb_tau1}
\end{equation}

\begin{figure}[t]
\begin{center}
\includegraphics[width=\figwidth]{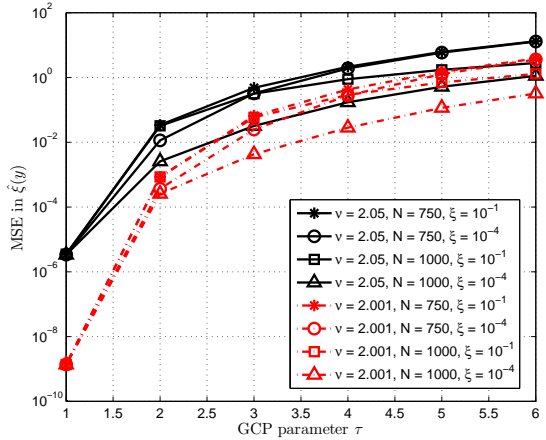}
\caption{Behavior of the MCRB \eqref{mcrb_gen} for the parametric form of the GCP, as a function of $\tau$, $\nu$, $N$ and noise variance $\xi$.}
\label{Nature_genMCRB_varytau}
\end{center}
\end{figure}

In Fig.~\ref{Nature_genMCRB_varytau}, we plot the expression in \eqref{mcrb_gen}. We observe that, in general, the bounds predict an increase in MSE for higher values
of $\tau$. Also, for given value of $N$,  the lower bounds at different signal to noise ratios (SNRs) converge as the value of $\tau$ increases, indicating that increasing $\tau$
renders the bound insensitive to the SNR. The lower bounds also predict a smaller value of MSE for a lower value of $\nu$.

Thus far, we have  presented the lower bounds on the MSE in estimating the unknown parameters of the SBL problem when the noise variance is known. In the next section, we extend
the results to the case of unknown noise variance. 

\section{SMV-SBL: Lower Bounds when $\sigma^2$ is Unknown}
\label{LB_signotknown}
Let us denote the  unknown noise variance as $\xi = \sigma^{2}$. In the Bayesian formulation, the noise variance is associated with a prior, and since the IG prior is
conjugate to the Gaussian likelihood $\pr_{\vecY|\vecX,\Xi}(\vecy|\vecx,\xi)$, it is assumed that $\sigma^2 \sim \mathcal{IG}(c,d)$ \cite{Sparse_RVM}, i.e., $\xi = \sigma^{2}$ is
distributed as
\begin{equation}
\pr_{\Xi}(\xi) \triangleq \frac{d^c}{\Gamma(c)}\xi^{(-c-1)}\exp\left\{-\frac{d}{\xi}\right\}; \quad \xi \in (0,\infty), ~ c,d > 0.
\label{IG_noisevar}
\end{equation}

Under this assumption, one can marginalize the unknown noise variance and obtain the likelihood $\pr(\vecy|\vecx)$ as
\begin{eqnarray}
&\pr(\vecy|\vecx) \triangleq \int_{\xi=0}^{\infty}\pr(\vecy, \xi|\vecx) \mathrm{d}\xi \nonumber\\
&= \frac{(2d)^c\Gamma\left(\frac{N}{2}+c\right)}{\Gamma(c)(\pi)^{N/2}}\left((\vecy -
\bmphi\vecx)^T(\vecy - \bmphi\vecx) + 2d\right)^{-\left(\frac{N}{2}+c\right)}\! ,
\end{eqnarray}
which is a multivariate Student-$t$ distribution.  It turns out that the straightforward approach of using the above multivariate likelihood to directly compute lower bounds for
the various cases given in the previous section is analytically intractable, and that the lower bounds cannot be computed in closed form. Hence, we compute lower bounds from the
\emph{joint} pdf, i.e., we derive the HCRB and BCRBs for the unknown vector $\bmtheta = [\vecx^T, \bmgm^T, \xi]^T$ with the MSE matrix $\mathbf{E}_{\xi}^{\bmtheta}$ defined by
\eqref{error_mat_gen}.\footnote{We use the subscript $\xi$ to indicate that the error matrices and bounds are obtained for the case of unknown noise variance.} Using the
assumptions and notation from the previous sections, we obtain the following proposition.

\begin{prop}
\label{prop3}
For the signal model in \eqref{signal_model}, the HCRB on the MSE matrix $\mathbf{E}_{\xi}^{\bmtheta}$  of the unknown vector $\bmtheta = [{\bmtheta'}^T,\xi]^T$, where
$\bmtheta' = [\vecx^T, \bmgm^T]^T$, with the distribution of the compressible vector $\vecx$ given by $\mathcal{N}(0,\bmup)$, where $\bmgm$ is modeled as a
deterministic or as a random parameter distributed as $\prod_{i = 1}^{L}\mathcal{IG}\left(\frac{\nu}{2}, \frac{\nu}{2\lambda}\right)$, and $\xi$ is modeled as a  deterministic parameter, is given by $(\mathbf{H}_{\xi}^{\bmtheta})^{-1}$, where
\begin{eqnarray}
\mathbf{H}_{\xi}^{\bmtheta} =
\begin{bmatrix}
\mathbf{H}^{\bmtheta'} &   \mathbf{0}_{L \times 1} \\
\mathbf{0}_{1 \times L}  & \frac{N}{2 \xi^2} 
\end{bmatrix}.
\label{prop4_mat}
\end{eqnarray}
\end{prop}
In the above expression, with a slight abuse of notation, $\mathbf{H}^{\bmtheta'}$ is the FIM given by \eqref{case1_HCRB} when $\bmgm$ is unknown deterministic and by \eqref{bcrb_gamma} when
$\bmgm$ is random.\\
\emph{Proof:} See Appendix \ref{pf_prop3}.\\
\indent The lower bound on the estimation of $\xi$ matches with known lower bounds on noise variance estimation (see Sec. 3.5 in \cite{Kay}). One disadvantage of such a
bound on $\hat{\xi}(\vecy)$ is that the knowledge of the noise variance is essential to compute the bound, and hence, it cannot be computed offline. Instead, assigning a hyperprior
to $\xi$ would result in a lower bound that only depends on the parameters of the hyperprior, which are assumed to be known, allowing the bound to be computed offline. We state the
following proposition in this context.
\begin{prop}
\label{prop4}
For the signal model in \eqref{signal_model}, the HCRB on the  MSE matrix $\mathbf{E}_{\xi}^{\bmtheta}$ of the unknown vector $\bmtheta = [{\bmtheta'}^T, \xi]^T$, where
$\bmtheta' = [\vecx^T, \bmgm^T]^T$, with the distribution of the vector $\vecx$ given by $\mathcal{N}(0,\bmup)$, where $\bmgm$ is modeled as a deterministic parameter or as a random parameter distributed as $\prod_{i = 1}^{L}\mathcal{IG}\left(\frac{\nu}{2}, \frac{\nu}{2\lambda}\right)$, and with the random parameter $\xi$
distributed as $\mathcal{IG}(c,d)$, is given by $(\mathbf{H}_{\xi}^{\bmtheta})^{-1}$, where
\begin{eqnarray}
\mathbf{H}_{\xi}^{\bmtheta} =
\begin{bmatrix}
\mathbf{H}^{\bmtheta'} &   \mathbf{0}_{L \times 1} \\
\mathbf{0}_{1 \times L}  & \frac{c(c+1)(N/2+c+3)}{d^2}
\end{bmatrix}.
\label{bcrb_unknoi}
\end{eqnarray}
\end{prop}
In \eqref{bcrb_unknoi}, $\mathbf{H}^{\bmtheta'}$ is the FIM given in \eqref{case1_HCRB} when $\bmgm$ is unknown deterministic and by \eqref{bcrb_gamma} when $\bmgm$ is random.\\
\emph{Proof:}
See Appendix \ref{pf_prop4}. 

In SBL problems, a non-informative prior on $\xi$ is typically preferred, i.e.,  the distribution of the noise variance is modeled to be as flat as possible. In \cite{Sparse_RVM},
it was observed that a non-informative prior is obtained when $c,d \rightarrow 0$. However, as $c,d \rightarrow 0$, the bound in \eqref{bcrb_unknoi} is
indeterminate. In Sec.~\ref{sim_res}, we illustrate the performance of the lower bound in \eqref{bcrb_unknoi} for practical values of $c$ and $d$.
\subsection{Marginalized Bounds}

In this subsection, we obtain lower bounds on  the MSE of the estimator $\hat{\xi}(\vecy)$, in the presence of nuisance variables in the joint distribution. To start with, we
consider the marginalized distributions of $\bmgm$ and $\xi$, i.e., $\pr_{\vecY; \bmgm, \xi}(\vecy;\bmgm, \xi)$ where both, $\bmgm$ and $\xi$ are deterministic variables. Since the
unknowns are deterministic, the regularity condition has to be satisfied for $\bmtheta = [\bmgm^T, \xi]^T$. We state the following theorem.

\begin{thm}
\label{thm4}
For the signal model in \eqref{signal_model},  the log likelihood function $\log\pr_{\vecY;\bmgm,\xi}(\vecy;\bmgm,\xi)$ satisfies the regularity condition \cite{Kay}. Further, the
MCRB on the MSE matrix $\mathbf{E}_{\xi}^{\bmtheta}$ of the unknown deterministic vector $\bmtheta = [\bmgm^T, \xi]^T$ is given by $\mathbf{E}_{\xi}^{\bmtheta} \succeq
({\mathbf{M}_{\xi}^{\bmtheta}})^{-1}$, where
\begin{equation}
\mathbf{M_{\xi}^{\bmtheta}} \triangleq \begin{bmatrix}
\mathbf{M}_{\xi}^{\bmtheta}(\bmgm) &   \mathbf{M}_{\xi}^{\bmtheta}(\bmgm,\xi)\\
\mathbf{M}_{\xi}^{\bmtheta}(\xi,\bmgm) & \mathbf{M}_{\xi}^{\bmtheta}(\xi)
\end{bmatrix},
\label{crlb_case3_unkn}
\end{equation}
where the $(ij)^{\text{th}}$ entry of the matrix  $\mathbf{M_{\xi}^{\bmtheta}}(\bmgm)$ is given by $(\mathbf{M_{\xi}^{\bmtheta}}(\bmgm))_{ij} = \frac{1}{2}\left\{(\Phi_j^T
\bmSigma_y^{-1}\Phi_i)^2\right\}$, and $\mathbf{M^{\bmtheta}_{\xi}}(\xi) = \frac{1}{2}\mbox{Tr}(\bmSigma_y^{-2})$. Further, 
$(\mathbf{M}_{\xi}^{\bmtheta}(\bmgm,\xi))_i = (\mathbf{M}_{\xi}^{\bmtheta}(\xi,\bmgm))_i =  \frac{\Phi_i^T\bmSigma_y^{-2}\Phi_i}{2}$, $i,j = 1, 2, \ldots, L$.
\end{thm}
\emph{Proof:} See Appendix \ref{pf_thm3}.

\emph{Remark:} From the graphical model in  Fig.~\ref{Graphical_model_system}, it can be seen that the branches consisting of $\gamma_i$ and $\xi$ are independent conditioned on $\vecx$.
However, when $\vecx$ is marginalized, the nodes $\xi$ and $\gamma_i$ are connected, and hence, Lemma~\ref{prop_gen_noise} is no longer valid. Due to this, the lower bound on $\bmgm$ depends on $\xi$
and vice versa, i.e., $\mathbf{M_{\xi}^{\bmtheta}(\bmgm)}$ and $\mathbf{M_{\xi}^{\bmtheta}(\xi)}$ depend on both $\xi$ and $\bmup = \mbox{diag}(\bmgm)$ through $\bmSigma_y = \xi
\mathbf{I}_{N \times N}+\bmphi \bmup \bmphi^T$.

Thus far, we have presented several bounds for the  MSE performance of the estimators $\hat{\vecx}(\vecy)$, $\hat{\bmgm}(\vecy)$ and $\hat{\xi}(\vecy)$ in the SMV-SBL framework. In
the next section, we derive Cram\'{e}r-Rao type lower bounds for the MMV-SBL signal model.

\section{Lower Bounds for the MMV-SBL}
\label{LB_MMVSBL}
In this section, we provide Cram\'{e}r-Rao type  lower bounds for the estimation of unknown parameters in the MMV-SBL model given in \eqref{mmv_systemmodel}. We consider the estimation of the compressible vector $\vecw$ from the vector of observations $\vect$, which contain the stacked columns of $\matW$ and $\matT$, respectively. In the MMV-SBL model,
each column of $\matW$ is distributed as $\vecw_i \sim \mathcal{N}(0,\bmup)$, for $i = 1, \hdots M$, and the likelihood is given by $\prod_{i = 1}^{M}\pr_{\matT|\matW_i,\Xi}(\vect_i|\vecw_i,\xi)$, where $\pr_{\matT|\matW_i\Xi}(\vect_i|\vecw_i,\xi) = \mathcal{N}(\bmphi\vecw_i,\xi)$ and $\xi = \sigma^2$. The modeling
assumptions on $\bmgm$ and $\xi$ are the same as in the SMV-SBL case, given by \eqref{IG_gamma} and \eqref{IG_noisevar}, respectively \cite{WipfRao_MMV}. 

Using the notation developed in Sec.~\ref{prelims},  we derive the bounds for the MMV SBL case similar to the SMV-SBL cases considered in Secs.~\ref{LB_sigknown} and~\ref{LB_signotknown}. Since the derivation of these bounds follow along the same lines as in the previous sections, we simply state results in Table~\ref{MMVSBL_bounds}.
\begin{table}[!ht]
\begin{center}
\begin{tabular}{|c|c|c|c|}
\hline
 \textbf{Bound Derived} & \textbf{Expression}\\ \hline \hline
 HCRB on $\hat{\bmgm}(\vecy)$ & $\mathbf{H}_M^{\bmtheta} = \mbox{diag}\left(\frac{M}{2 \gamma_i^2}\right)$, $i = 1, 2 \hdots, L$  \\ \hline
 BCRB on $\hat{\bmgm}(\vecy)$ & $\mathbf{B}_M^{\bmtheta} = \frac{\lambda^2(\nu+2)(M+\nu+6)}{2 \nu}\mathbf{I}_{L \times L} $\\ \hline
 MCRB on $\hat{\bmgm}(\vecy)$ & $\mathbf{M}_M^{\bmtheta} =  [\mathbf{M}_{ij}^{\bmtheta}]$,  \\
&  where $\mathbf{M}_{ij}^{\bmtheta} = \frac{M}{2}(\Phi_j^T \bmSigma_y^{-1}\Phi_i)^2$\\ \hline
 HCRB on $\hat{\vecw}(\vecy)$ & $\mathbf{H}_M^{\bmtheta} = \left(\frac{\bmphi^T \bmphi}{\sigma^2}  + {\bmup^{-1}}\right)\otimes \mathbf{I}_{M \times M}$ \\ \hline
 BCRB on $\hat{\vecw}(\vecy)$ & $\mathbf{B}_M^{\bmtheta} = \left(\frac{\bmphi^T \bmphi}{\sigma^2}  + {\lambda}\mathbf{I}_{L \times L}\right)\otimes \mathbf{I}_{M \times M}$
\\ \hline
 HCRB on $\hat{\xi}(\vecy)$ & $\mathbf{H}_{M,\xi}^{\bmtheta} = \left(\frac{MN}{2 \xi^2}\right)$\\ \hline
 BCRB on $\hat{\xi}(\vecy)$ & $\mathbf{B}_{M,\xi}^{\bmtheta} = \frac{c \left(\frac{MN}{2}+c+3\right)(c+1)}{d^2}$\\ \hline
 MCRB on $[\hat{\bmgm}(\vecy)^T$, $\hat{\xi}(\vecy)]^T$ & $\mathbf{M}_{M,\xi}^{\bmtheta} = {M}\times \mathbf{M_{\xi}^{\bmtheta}}$ \\ \hline
\end{tabular}
\end{center}
\caption{Cram\'{e}r-Rao Type Bounds for the MMV-SBL Case.}
\label{MMVSBL_bounds} 
\end{table}

We see that the lower bounds on $\hat{\bmgm}(\vecy)$ and $\hat{\xi}(\vecy)$ are reduced by a factor of $M$ compared to the SMV case, which is intuitively satisfying. 
It turns out that it is not possible to obtain the MCRB on $\vecw$ in the MMV-SBL setting, since closed form expressions for the FIM are not available.

In the next section, we consider two popular algorithms for SBL and graphically illustrate the utility of the lower bounds.
\section{Simulations and Discussion}
\label{sim_res}

The vector estimation problem in the SBL  framework typically involves the joint estimation of the hyperparameter and the unknown compressible vector $\vecx$. Since the
hyperparameter estimation problem cannot be solved in closed form, iterative estimators are employed \cite{Sparse_RVM}. In this section, we consider the iterative updates based on
the EM algorithm first proposed in \cite{Sparse_RVM}. We also consider the algorithm proposed in \cite{wipfRao_ARD} based on the Automatic Relevance Determination (ARD) framework. We plot the MSE performance in estimating $\vecx$, $\bmgm$ and $\xi$ with the linear model in \eqref{signal_model} and
\eqref{mmv_systemmodel}, for the EM
algorithm, labeled \texttt{EM}, and the ARD based Reweighted $\ell_1$ algorithm, labeled~\texttt{ARD-SBL}. We compare the performance of the estimators against the derived lower bounds. 

\begin{figure}[t]
\begin{center}
\includegraphics[width=\figwidth]{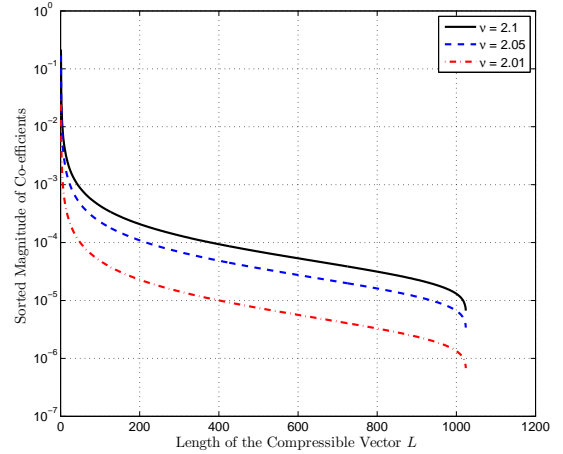}
\caption{Decay profile of the sorted magnitudes of \emph{i.i.d.} samples drawn from a Student-$t$ distribution.}
\label{sorted_magnitudes}
\end{center}
\end{figure}

We simulate the lower bounds for a  random underdetermined ($N < L$) measurement matrix $\bmphi$, whose entries are \emph{i.i.d.} and standard Bernoulli
$\left(\left\{+1,-1\right\}\right)$ distributed. A compressible signal of dimension $L$ is generated by sampling from a Student-$t$ distribution with the value of $\nu$ ranging
from $2.01$ to $2.05$, which is the range in which the signal is \emph{``somewhat''} compressible, for high dimensional signals~\cite{gribonval_compressible}.
Figure~\ref{sorted_magnitudes} shows the decay profile of the sorted magnitudes of $L = 1024$ \emph{i.i.d.} samples drawn from a Student-$t$ distribution for different $\nu$ and with the value of $\mathbb{E}(x_i^2)$ fixed at~$10^{-3}$.

\subsection{Lower Bounds on the MSE Performance of $\hat{\vecx}(\vecy)$}

\begin{figure}[t]
\begin{center}
\includegraphics[width=\figwidth]{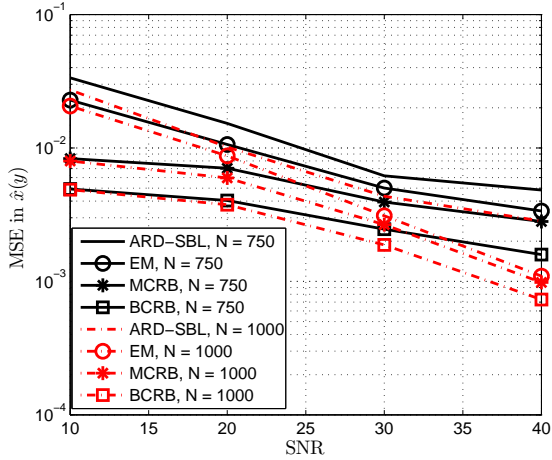}
\caption{The MSE performance of $\hat{\vecx}(\vecy)$ and the corresponding MCRB and BCRB, as a function of SNR, with $\nu = 2.01$.}
\label{MCRB_BCRB_MSE_varysnrN_p205}
\end{center}
\end{figure}

In this subsection, we compare the  MSE performance of the ARD-SBL estimator and the EM based estimator $\hat{\vecx}(\vecy)$. Figure~\ref{MCRB_BCRB_MSE_varysnrN_p205} depicts the
MSE performance of $\hat{\vecx}(\vecy)$ for different SNRs and $N = 750$ and $1000$, with $\nu = 2.01$. We compare it with the HCRB/BCRB derived in \eqref{case1_HCRB}, which is obtained
by assuming the knowledge of the realization of the hyperparameters $\bmgm$. We see that the MCRB derived in \eqref{mcrb_case4} is a tight lower bound on the MSE performance at
high SNR and $N$. 
\begin{figure}[t]
\begin{center}
\includegraphics[width=\figwidth]{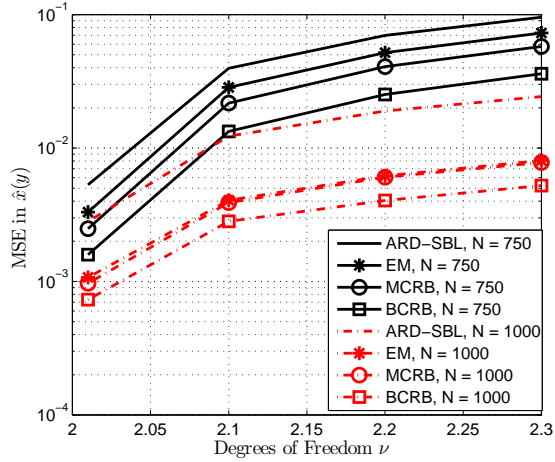}
\caption{The MSE performance of $\hat{\vecx}(\vecy)$ and the corresponding MCRB and BCRB, as a function of $\nu$, with SNR~=~$40$~dB.}
\label{MSE_MCRB_BCRB_varynu}
\end{center}
\end{figure}

\begin{figure}[t]
\begin{center}
\includegraphics[width=\figwidth]{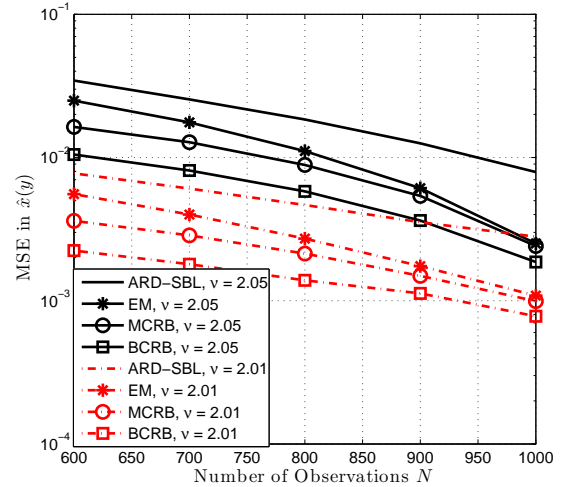}
\caption{The MSE performance of $\hat{\vecx}(\vecy)$ and the corresponding MCRB and BCRB, as a function of $N$, with SNR~=~$40$~dB.}
\label{MSE_MCRB_BCRB_varyN}
\end{center}
\end{figure}

Figure \ref{MSE_MCRB_BCRB_varynu}  shows the comparative MSE performance of the ARD-SBL estimator and EM based estimator as a function of varying degrees of freedom $\nu$, at an
SNR of $40$~dB and $N = 1000$ and $750$. As expected, the MSE performance of the algorithms is better at low values of $\nu$ since the signal is more compressible, and the MCRB and
BCRB also reflect this behavior. The MCRB is a tight lower bound, especially for high values of $N$. Figure~\ref{MSE_MCRB_BCRB_varyN} shows the MSE performance of
the ARD-SBL estimator and EM based estimator as a function of $N$, at an SNR of $40$~dB and for two different values of~$\nu$. The MSE performance of the EM
algorithm converges to that of the MCRB at higher~$N$.

\subsection{Lower Bounds on the MSE Performance of $\hat{\bmgm}(\vecy)$}
\begin{figure}[t]
\begin{center}
\includegraphics[width=\figwidth]{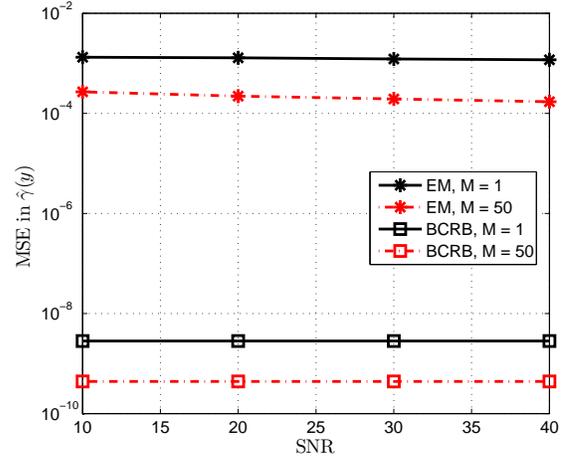}
\caption{The MSE performance of $\hat{\bmgm}(\vecy)$ and the corresponding HCRB, as a function of SNR, with $N = 1000$.}
\label{HCRB_EM_gamma}
\end{center}
\end{figure}

In this subsection, we compare the  different lower bounds for the MSE of the estimator $\hat{\bmgm}(\vecy)$ for the SMV and MMV-SBL system model. Figure~\ref{HCRB_EM_gamma} shows
the MSE performance of $\hat{\bmgm}(\vecy)$ as a function of SNR and $M$, when $\gamma$ is a random parameter, $N = 1000$ and $\nu = 2.01$. In this case, it turns out that there is a
large gap between the performance of the EM based estimate and the lower bound. 

When $\bmgm$ is deterministic, we first  note that the EM based ML estimator for $\bmgm$ is asymptotically optimal and the lower bounds are practical for large data samples
\cite{Kay}. The results are listed in Table \ref{gamma_bounds}. We see that for $L = 2048$ and $N = 1500$, the MCRB and BCRB are tight lower bounds, with MCRB being marginally
tighter than the BCRB. However, as $M$ increases, the gap between the MSE and the lower bounds increases.

\begin{table}[!ht]
\begin{center}
\begin{tabular}{|c|c|c|c|c|c|}
\hline
SNR(dB) & & $10$ & $20$ & $30$ & $40$ \\ \hline \hline
\multirow{3}{1.2cm}{$M = 1$}& MSE & $0.054$ &  $0.053$ &   $0.051$ & $0.050$ \\ \cline{2-6}
& MCRB & $0.052$ & $0.051$ &  $0.050$ & $0.049$ \\  \cline{2-6}
& BCRB & $0.049$ & $0.049$ &  $0.049$ & $0.049$ \\ \cline{1-6} 
\multirow{3}{1.5cm}{$M = 50$} & MSE  & $0.0450$ & $0.039$ & $0.035$ & $0.030$\\ \cline{2-6}
& MCRB $\times10^{-2}$& $0.12$ & $0.11$ &$0.10$ &$0.09$\\ \cline{2-6}
& BCRB$\times10^{-3}$ & $0.977$ & $0.977$ & $0.977$ & $0.977$\\ \hline
\end{tabular}
\end{center}
\caption{Values of the MSE of the estimator $\hat{\bmgm}(\vecy)$, the MCRB and the BCRB, for $\bmtheta_d = [\bmgm]$ as a function of SNR, for $N = 1500$.}
\label{gamma_bounds} 
\end{table}

\subsection{Lower Bounds on the MSE Performance of $\hat{\xi}(\vecy)$}

\begin{figure}[t]
\begin{center}
\includegraphics[width=\figwidth]{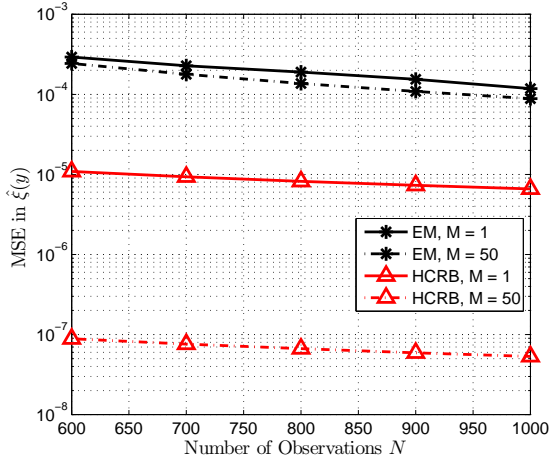}
\caption{The MSE performance of $\hat{\xi}(\vecy)$ and its HCRB, as a function of~$N$.}
\label{HCRB_unknoi_random}
\end{center}
\end{figure}

In Fig.~\ref{HCRB_unknoi_random}, we compare the lower bounds on the MSE of the estimator $\hat{\xi}(\vecy)$ in the SMV and MMV-SBL settings, for different values of $N$ and $M$. Here, $\xi$ is sampled from the IG
pdf \eqref{IG_noisevar}, with parameters $c = 3$ and~$d = 0.2$. 


When $\xi$ is deterministic, the EM based ML  estimator for $\bmgm$ is asymptotically optimal and the lower bounds are practical for large data samples \cite{Kay}.
Table~\ref{detxi_bounds} lists the MSE values of $\hat{\xi}(\vecy)$, the corresponding HCRB and MCRB for deterministic but unknown noise variance, while the true noise variance is
fixed at $10^{-3}$. We see that for $L = 2048$ and $N = 1500$, the MCRB is marginally tighter than the HCRB. However, when the noise
variance is random, we see from Fig.~\ref{HCRB_unknoi_random} that there is a large gap between the MSE performance and the HCRB.

\begin{table}[!ht]
\begin{center}
\begin{tabular}{|c|c|c|c|c|c|}
\hline
& $N$ & $1500$ & $1600$ & $1700$ & $1800$ \\ \hline \hline
\multirow{3}{1cm}{$M = 1$}& MSE $\times 10^{-8}$ & $0.736$  &  $0.663$ &   $0.636$ &
$0.592$ \\ \cline{2-6}
& MCRB$\times 10^{-8}$ & $0.380$ & $0.340$ &  $0.307$ & $0.279$ \\  \cline{2-6}
& HCRB$\times 10^{-8}$ & $0.133$ & $0.125$ &  $0.118$ & $0.111$ \\ \cline{1-6} 
 \multirow{3}{1.5cm}{$M = 50$} & MSE $\times10^{-9}$& $0.930$ & $0.892$ & $0.866$ & $0.847$\\ \cline{2-6}
& MCRB$\times 10^{-10}$ & $ 0.680$ & $0.652$ & $0.614$ & $0.573$\\ \cline{2-6}
& HCRB$\times 10^{-10}$ & $0.267$ & $0.250$ & $0.235$ & $0.222$\\ \hline
\end{tabular}
\end{center}
\caption{Values of the MSE of the estimator $\hat{\xi}(\vecy)$, the MCRB and the HCRB for $\theta_d = [\xi]$, as a function of $N$.}
\label{detxi_bounds} 
\end{table}

\section{Conclusion}
\label{sec:concl}
In this work, we derived Cram\'{e}r-Rao type lower bounds on the MSE, namely,  the HCRB, BCRB and MCRB, for the SMV-SBL and the MMV-SBL problem of estimating compressible signals.
We used a hierarchical model for the compressible priors to obtain the bounds under various assumptions on the unknown parameters. The bounds derived by assuming
a hyperprior distribution on the hyperparameters themselves provided key insights into the MSE performance of SBL and the values of the parameters that govern these hyperpriors. We derived
the MCRB for the generalized compressible prior distribution, of which the Student-$t$ and Generalized Pareto prior distribution are special cases. We showed that the MCRB is tighter than the BCRB.
We compared the lower bounds with the MSE performance of the ARD-SBL and the EM algorithm using Monte Carlo simulations. The numerical results illustrated the  near-optimality of EM based updates for SBL, which makes it attractive for practical implementations. 

\appendix

\subsection{Proof of Proposition \ref{prop1}}
\label{pf_prop1}
Using the graphical model of Fig.~\ref{Graphical_model_system} in \eqref{Infomat_def_gen}, 
\begin{eqnarray}
\mathbf{H}^{\bmtheta}(\vecx) &\triangleq& -\mathbb{E}_{\vecY, \vecX;\bmgm}\left[ \nabla_\vecx^2 \log \pr_{\vecY, \vecX; \bmgm}(\vecy, \vecx;\bmgm)\right]\nonumber\\
&=& -\mathbb{E}_{\vecY, \vecX;\bmgm}\left[ \nabla_\vecx \left(\frac{\bmphi^T (\vecy - \bmphi\vecx)}{\sigma^2} - {\bmup^{-1} \vecx}\right) \right] \nonumber\\
&=&  \frac{\bmphi^T
\bmphi}{\sigma^2}  + \bmup^{-1}.
\label{bound_likelihood}
\end{eqnarray}
Similarly, it is straightforward to show that $\nabla_{\vecx} \nabla_{\bmgm}\log \pr_{\vecY, \vecX; \bmgm}(\vecy,\vecx; \bmgm) = \textnormal{diag}\left(\frac{ x_1}{\gamma_1^2}, ~
\frac{ x_2}{\gamma_2^2},~ \hdots,~ \frac{ x_L}{\gamma_L^2}\right)$. Since $x_i$ are zero mean random variables,
\begin{eqnarray}
\mathbf{H}^{\bmtheta}(\bmgm, \vecx) = -\mathbb{E}_{\vecY, \vecX;\bmgm}\left[\nabla_{\bmgm} \nabla_{\vecx}\log \pr_{\vecY,\vecX;\bmgm}(\vecy,\vecx; \bmgm) \right] = \mathbf{0}_{L
\times L}, \nonumber
\end{eqnarray}
\begin{equation}
\mathbf{H}^{\bmtheta}(\bmgm) =  -\mathbb{E}_{\vecY,\vecX;\bmgm} \left[\nabla_{\bmgm}^2 (\log \pr_{\vecY|\vecX}(\vecy|\vecx) + \log \pr_{\vecX; \bmgm}(\vecx ;
\bmgm))\right]. \nonumber
\end{equation}
Now, since $\log \pr_{\vecX;\bmgm}(\vecx;\bmgm) = \sum_{i = 1}^{L}\log \pr_{\vecX;\bmgm}(\vecx_i;\bmgm_i)$, we get,
\begin{eqnarray}
\frac{\partial^2 \log \pr_{\vecX; \bmgm}(\vecx; \bmgm)}{\partial \gamma_i\partial \gamma_j} &=&  \left\{ \begin{array}{ll}
\frac{1}{2 \gamma_i^2} - \frac{x_i^2}{\gamma_i^3} & \mbox{if}~ i = j\\
0 & \mbox{if}~ i \neq j.
\end{array} \right.
\end{eqnarray}
Taking $-\mathbb{E}_{\vecX;\bmgm}(\cdot)$ on both sides of the above equation and noting that $\mathbb{E}_{\vecX;\bmgm}(x_i^2) = \gamma_i$, we obtain
\begin{eqnarray}
\mathbf{H}^{\bmtheta}(\bmgm)  &=& \mbox{diag}\left(-\mathbb{E}_{\vecX;\bmgm}\left[\frac{\partial^2 \log \pr_{\vecX; \bmgm}(\vecx; \bmgm)}{\partial \gamma_i^2}\right]\right)
\nonumber\\
&=& \mbox{diag}\left(\left[\frac{1}{2\gamma_1^2}, \hdots, \frac{1}{2\gamma_L^2}\right]\right).
\end{eqnarray}
This completes the proof.

\subsection{Proof of Proposition \ref{prop2}} \label{prop2proof}

Using the graphical model of Fig.~\ref{Graphical_model_system} in \eqref{Infomat_def_gen}, 
\begin{eqnarray}
\mathbf{B}^{\bmtheta}(\vecx) &\triangleq& -\mathbb{E}_{\vecY, \vecX,\Gamma}\left[ \nabla_\vecx^2 \log \pr_{\vecY, \vecX, \Gamma}(\vecy, \vecx;\bmgm)\right]\nonumber\\
&=& -\mathbb{E}_{\vecY, \vecX, \Gamma}\left[ \nabla_\vecx \left(\frac{\bmphi^T (\vecy - \bmphi\vecx)}{\sigma^2} - {\bmup^{-1} \vecx}\right) \right]\nonumber\\
&=& \mathbb{E}_{\Gamma}\left[\frac{\bmphi^T \bmphi}{\sigma^2}  + \bmup^{-1}\right]\\
&=& \frac{\bmphi^T \bmphi}{\sigma^2}+\mathbb{E}_{\Gamma}\left[ \bmup^{-1}\right].
\end{eqnarray}
The expression for $\mathbb{E}_{\Gamma}\left[ \bmup^{-1}\right]$ w.r.t. $\gamma_i$ is given by,
 \begin{eqnarray}
\mathbb{E}_{\Gamma}\left[ \frac{1}{\gamma_i}\right] &=& K_{\gamma}\int_{\gamma_i = 0}^{\infty}\gamma_i^{\left(-\frac{\nu}{2} -
2\right)}\exp\left\{-\frac{\nu}{2\lambda\gamma_i}\right\} \mathrm{d}\gamma_i\\
&=& K_{\gamma} \frac{\Gamma\left(\frac{\nu}{2}+1\right)}{\left(\frac{\nu}{2\lambda}\right)^{\frac{\nu}{2}+1}}\underbrace{\int_{\gamma_i =
0}^{\infty}\mathcal{IG}\left(\frac{\nu}{2}+1,\frac{\nu}{2\lambda}\right) \mathrm{d}\gamma_i}_{=1} \nonumber\\
&=& \lambda,
\end{eqnarray}
since $K_{\gamma} = \left(\frac{\nu}{2\lambda}\right)^{\nu/2}\left(\Gamma\left(\frac{\nu}{2}\right)\right)^{-1}$. Hence, the overall bound is given by
\begin{equation}
\mathbf{B}^{\bmtheta}(\vecx) =  \frac{\bmphi^T \bmphi}{\sigma^2}+\lambda \mathbf{I}_{L \times L}.
\end{equation}

Using the graphical model of Fig.~\ref{Graphical_model_system} in
\eqref{Infomat_def_gen}, for $\bmtheta = [\vecx^T, \bmgm^T]^T$, $\mathbf{B}^{\bmtheta}(\bmgm)$ is defined  as
\begin{eqnarray}
&\mathbf{B}^{\bmtheta}(\bmgm) \triangleq  -\mathbb{E}_{\vecY,\vecX,\Gamma} \left[\nabla_{\bmgm}^2 \left(\log \pr_{\vecY|\vecX}(\vecy|\vecx)\right.\right.\nonumber\\
&\left.\left.+ \log \pr_{\vecX|\Gamma}(\vecx|\bmgm) + \log \pr_{\Gamma}(\bmgm)\right)\right].
\end{eqnarray}
Since the expressions  for $\log \pr_{\vecX|\Gamma}(\vecx|\bmgm) $ and $ \log \pr_{\Gamma}(\bmgm)$ are separable and symmetric w.r.t. $\gamma_i$, the off-diagonal terms of
$\mathbf{B}^{\bmtheta}(\bmgm)$ are zero, and it is sufficient to evaluate the diagonal terms              $ -\mathbb{E}_{\vecY, \vecX, \Gamma}\left(\frac{\partial^2 (\log
\pr_{\vecX|\Gamma}(\vecx|\bmgm) + \log \pr_{\Gamma}(\bmgm))}{\partial \gamma_i^2}\right)$. Differentiating the expression w.r.t. $\gamma_i$ twice,
\begin{equation}
\frac{\partial^2\left( \log \pr_{\vecX|\Gamma}(\vecx| \bmgm) + \log \pr_{\Gamma}(\bmgm) \right)}{\partial \gamma_i^2} =  -\frac{(\nu + 1)}{2\gamma_i^2}+\frac{\nu}{\lambda
\gamma_i^3}.
\end{equation}
The expression for $-\mathbb{E}_{\Gamma}\left[-\frac{(\nu + 1)}{2\gamma_i^2}+\frac{\nu}{\lambda \gamma_i^3}\right]$ is given by
\begin{eqnarray}
&\mathbb{E}_{\Gamma}\left[\frac{(\nu + 1)}{2\gamma_i^2}-\frac{\nu}{\lambda \gamma_i^3}\right]= K_{\gamma} \nonumber\\
&   \int\limits_{\gamma_i = 0}^{\infty}\left[\frac{(\nu + 1){\gamma_i^{-2}}}{2}
-\frac{\nu{\gamma_i^{-3}}}{\lambda}\right]\gamma_i^{\left(-\frac{\nu}{2} - 1\right)}\exp\{-\frac{\nu}{2\lambda\gamma_i}\} \mathrm{d}\gamma_i,
\end{eqnarray}
where $K_{\gamma} = \left(\frac{\nu}{2\lambda}\right)^{\nu/2}\left(\Gamma\left(\frac{\nu}{2}\right)\right)^{-1}$. After some manipulation, it can be shown that the above integral reduces to
\begin{equation}
-\mathbb{E}_{\Gamma}\left[-\frac{(\nu + 1)}{2\gamma_i^2}+\frac{\nu}{\lambda \gamma_i^3}\right] = \frac{\lambda^2(\nu +2)(\nu + 7)}{2\nu}.
\end{equation}
Thus, the $(ij)^{\text{th}}$ component of $\mathbf{B}^{\bmtheta}(\bmgm, \vecx)$ is given by
\begin{equation}
(\mathbf{B}^{\bmtheta}(\bmgm, \vecx))_{ij} = \frac{\partial^2 \log \pr_{\vecX|\Gamma}(\vecx|\bmgm)}{\partial \gamma_i \partial x_i} = -\frac{x_i}{\gamma_i^2},
\end{equation}
and $\mathbf{B}^{\bmtheta}(\vecx, \bmgm) = (\mathbf{B}^{\bmtheta}(\bmgm, \vecx))^T$. Since $\mathbb{E}_{\vecX|\Gamma}(x_i) = 0$,  $\mathbf{B}^{\bmtheta}(\bmgm, \vecx)  = 
\mathbf{0}_{L \times L}$. This completes the proof.

\subsection{Proof of Theorem \ref{thm1}}
\label{pf_thm1}

To establish the regularity condition, the first order derivative  of the log likelihood $\log\pr_{\vecY;\bmgm}(\vecy;\bmgm)$ is required. This, in turn, requires the evaluation of
$\frac{\partial \log |\bmSigma_y|}{\partial \gamma_j}$ and $\frac{\partial \vecy^T\bmSigma_y^{-1} \vecy}{\partial \gamma_j}$. Using the chain rule for differentiation~\cite{mat_cookbook}, we have
\begin{align}
\frac{\partial \log |\bmSigma_y|}{\partial \gamma_j} &=  \mbox{Tr}\left\{\left(\frac{\partial\log |\bmSigma_y|}{\partial \bmSigma_y}\right)^T\frac{\partial \bmSigma_y}{\partial
\gamma_j}  \right\}\nonumber\\
 &= \mbox{Tr}\left\{(\bmSigma_y^{-1})^T\Phi_j\Phi_j^T \right\} = \Phi_j^T \bmSigma_y^{-1} \Phi_j.
\end{align}
Here, we have used the identity $\nabla_{X} \log |X| =  X^{-1}$ \cite{mat_cookbook} and results from vector calculus \cite{mat_cookbook} to obtain $\frac{\partial
\bmSigma_y}{\partial \gamma_j} = \Phi_j\Phi_j^T$, where $\Phi_j$ is the $j^{\text{th}}$ column of $\bmphi$. 
Similarly, the derivative of $\vecy^T\bmSigma_y^{-1} \vecy$ can be obtained as 
\begin{eqnarray}
\frac{\partial \vecy^T\bmSigma_y^{-1} \vecy}{\partial \gamma_j}  &=  \mbox{Tr} \left\{ \left(\frac{\partial \vecy^T\bmSigma_y^{-1} \vecy}{\partial \bmSigma_y^{-1}}\right)^T
\frac{\partial \bmSigma_y^{-1}}{\partial \gamma_j} \right \} \nonumber\\
&= -\Phi_j^T\bmSigma_y^{-1}\vecy\vecy^T\bmSigma_y^{-1}\Phi_j,
\end{eqnarray}
and hence,
\begin{equation}
\frac{\partial}{\partial \gamma_j}\log \pr_{\vecY;\bmgm}(\vecy;\bmgm) =  \frac{\Phi_j^T\bmSigma_y^{-1}\vecy\vecy^T\bmSigma_y^{-1}\Phi_j - \Phi_j^T \bmSigma_y^{-1} \Phi_j}{2}.
\label{firstdiff_margam}
\end{equation}
Taking $\mathbb{E}_{\vecY;\bmgm}(\cdot)$ on both the sides of the above equation, 
\begin{multline}
\mathbb{E}_{\vecY;\bmgm}\left[\frac{\partial}{\partial \gamma_j} \log \pr_{\vecY; \bmgm}(\vecy;\bmgm)\right] \\
= \frac{\Phi_j^T\bmSigma_y^{-1}\left\{\mathbb{E}_{\vecY;\bmgm}(\vecy\vecy^T)\right\}\bmSigma_y^{-1}\Phi_j - \Phi_j^T \bmSigma_y^{-1} \Phi_j 
}{2}= 0,
\label{reg_pf}
\end{multline}
since $\mathbb{E}_{\vecY}(\vecy\vecy^T) = \bmSigma_y$. Hence, the pdf satisfies the required regularity constraint.

Now, the MCRB for $\bmtheta = [\bmgm]$ is obtained by computing the second derivative of the log likelihood, as follows:
\begin{multline}
-\frac{\partial^2 }{\partial \gamma_i \partial \gamma_j} \log \pr_{\vecY,\bmgm}(\vecy;\bmgm) \\
=  \frac{1}{2}\frac{\partial}{\partial \gamma_i}(\Phi_j^T \bmSigma_y^{-1} \Phi_j
-(\Phi_j^T\bmSigma_y^{-1}\vecy)^2)\\
 = \frac{1}{2}\mbox{Tr}\left\{ \Phi_j \Phi_j^T (-\bmSigma_y^{-1}\Phi_i\Phi_i^T\bmSigma_y^{-1})\right\}\\
- (\Phi_j^T\bmSigma_y^{-1}\vecy)\mbox{Tr}\left\{\left(\frac{\partial (\Phi_j^T\bmSigma_y^{-1}\vecy)}{\partial \bmSigma_y^{-1}} \right)^{T} \frac{\partial
\bmSigma_y^{-1}}{\partial \gamma_i}\right\}\\
= -\frac{1}{2}\left(\Phi_j^T \bmSigma_y^{-1}\Phi_i\right) \left(\Phi_i^T\bmSigma_y^{-1}\Phi_j\right) \\
+ \left(\Phi_j^T\bmSigma_y^{-1}\vecy\right)\left(\vecy^T\bmSigma_y^{-1}\Phi_i\right)\left(\Phi_i^T\bmSigma_y^{-1}\Phi_j\right).
\end{multline}
Taking $-\mathbb{E}_{\vecY;\bmgm}(\cdot)$ on both the sides of the above expression, 
\begin{eqnarray}
(\mathbf{M^{\bmgm}})_{ij} \triangleq  -\mathbb{E}_{\vecY;\bmgm}\left[\frac{\partial^2 \log \pr_{\vecY;\bmgm}(\vecy;\bmgm)}{\partial \gamma_i \partial\gamma_j}\right] = \frac{(\Phi_j^T
\bmSigma_y^{-1}\Phi_i)^2}{2},
\label{MCRB_detgm}
\end{eqnarray}
as stated in \eqref{crlb_case3}. This completes the proof.
\subsection{Proof of Theorem \ref{thm2}}
\label{pf_thm2}
The proof follows from the proof for Theorem \ref{thm3} in Appendix \ref{pf_thm4} by substituting $\tau = 2$.
\subsection{Proof of Theorem \ref{thm3}}
\label{pf_thm4}

The MCRB for estimation of the compressible random vector with $\bmtheta = [\vecx]$ is given by
\begin{multline}
\mathbf{M}^{\vecx}=-\mathbb{E}_{\vecY,\vecX}[\nabla^2_\vecx\log \pr_{\vecY, \vecX}(\vecy,\vecx)] \\
 = -\mathbb{E}_{\vecY,\vecX}[\nabla^2_\vecx\log \pr_{\vecY|\vecX}(\vecy|\vecx) +
\nabla^2_\vecx\log \pr_{\vecX}(\vecx)].
\end{multline}
The first term above is given by
\begin{eqnarray}
&-\mathbb{E}_{\vecY, \vecX} \left[\nabla^2_\vecx\log \pr_{\vecY|\vecX}(\vecy|\vecx)\right] = -\mathbb{E}_{\vecY, \vecX}\left[ \nabla_\vecx \frac{\bmphi^T (\vecy -
\bmphi\vecx)}{\sigma^2} \right]\nonumber\\
&= -\mathbb{E}_{\vecY, \vecX} \left[\frac{-\bmphi^T \bmphi}{\sigma^2}  \right] = \frac{\bmphi^T \bmphi}{\sigma^2}.
\label{mcrbx_firstpart}
\end{eqnarray}

Note that $\pr_{\vecX}(\vecx)$ is not differentiable if any of its components $x_i = 0$. However, the measure of $x_i = 0$ is zero since the distribution  is continuous, and hence,
this condition can be safely ignored. Now, 
\begin{eqnarray}
\frac{\partial  \log \pr_{\vecX}(\vecx) }{\partial x_i}= \left\{ \begin{array}{ll}
-\frac{(\nu+1)\lambda x_i^{\tau - 1}}{\left(\nu+{\lambda x_i^{\tau}}\right)} & \mbox{if} ~~ x_i>0\\
(-1)^{\tau}\frac{(\nu+1)\lambda x_i^{\tau - 1}}{\left(\nu+(-1)^{\tau}{\lambda x_i^{\tau}}\right)} & \mbox{if}~~ x_i<0.
\end{array} \right. \nonumber
\end{eqnarray}
First, we consider the case of $x_i >0$. Differentiating the above w.r.t. $x_i$ again, we obtain
\begin{multline}
\frac{\partial^2}{\partial x_i^2}\log\pr_{\vecX}(\vecx) = \frac{-(\nu+1)\lambda(\tau - 1) x_i^{\tau - 2}}{\left(\nu+{\lambda x_i^\tau}\right)}  \\
+
\frac{\lambda^2\tau(\nu+1) x_i^{2\tau - 2}}{\left(\nu+{\lambda x_i^\tau}\right)^2}.
\end{multline}
Taking $-\mathbb{E}_{\vecX}(\cdot)$ on both sides of the above equation, we get
\begin{multline}
-\mathbb{E}_{\vecX}\left(\frac{\partial^2}{\partial x_i^2}\log\pr_{\vecX}(\vecx)\right) = \frac{K(\nu+1)\lambda}{\nu}\\
\int_0^{\infty} \left( \frac{(\tau - 1) x_i^{\tau - 2}}{\left(1+\frac{\lambda x_i^\tau}{\nu}\right)^{\frac{\nu+\tau+1}{\tau}}} - \frac{\lambda\tau x_i^{2\tau -
2}}{\nu\left(1+\frac{\lambda x_i^\tau}{\nu}\right)^{\frac{\nu+2\tau+1}{\tau}}} \right) \mathrm{d}x_i.
\label{com_integral}
\end{multline}
The above can be simplified using the transformation  $t_i = \frac{\lambda x_i^\tau}{\nu}$ and using $\int_{0}^{\infty} \frac{t^{u-1}}{(1+t)^{u+v}}\mathrm{d}t =
\frac{\Gamma(u)\Gamma(v)}{\Gamma(u+v)}$, we get
\begin{align}
&-\mathbb{E}_{\vecX}\left(\frac{\partial^2}{\partial x_i^2}\log\pr_{\vecX}(\vecx)\right) = \frac{K(\nu+1)(\tau -
1)}{\tau}\left(\frac{\lambda}{\nu}\right)^{1/\tau}\nonumber\\
&\Gamma\left(1-\frac{1}{\tau}\right)\left\{\frac{\Gamma\left(\frac{\nu+\tau+2}{\tau}\right) -
\frac{1}{\tau}\Gamma\left(\frac{\nu+2}{\tau}\right)}{\Gamma\left(\frac{v+2\tau+1}{\tau}\right)} \right\}\quad \mbox{for} \quad x_i > 0. 
\end{align}
For the case of $x_i < 0$ also, the expression reduces to the integral given in \eqref{com_integral}. Hence, we have 
\begin{multline}
-\mathbb{E}_{\vecX}\left(\frac{\partial^2}{\partial x_i^2}\log\pr_{\vecX}(\vecx)\right)  = \frac{K(\nu+1)^2(\tau - 1)}{\tau (\nu + \tau +
1)}\left(\frac{\lambda}{\nu}\right)^{1/\tau}\\
\left( \frac{\Gamma\left(\frac{\tau-1}{\tau}\right)\Gamma\left(\frac{\nu+2}{\tau}\right)}{\Gamma\left(\frac{v+\tau+1}{\tau}
\right)}\right).
\end{multline}
Substituting the expression for $K$ in the above, we get
\begin{multline}
-\mathbb{E}_{\vecX}\left(\frac{\partial^2}{\partial x_i^2}\log\pr_{\vecX}(\vecx)\right)  = \frac{\tau^2(\nu+1)}{(\nu + \tau + 1)}
\left(\frac{\lambda}{\nu}\right)^{2/\tau}\\
\frac{\Gamma\left(\frac{\nu+2}{\tau}\right)\Gamma\left(2-\frac{1}{\tau}\right)}{\Gamma\left(\frac{1}{\tau}
\right)\Gamma\left(\frac { v } { \tau }\right)}.
\end{multline}
Combining the expression above and \eqref{mcrbx_firstpart}, we obtain the MCRB in~\eqref{mcrb_tau1}.
\subsection{Proof of Proposition \ref{prop3}}
\label{pf_prop3}
In this case, we define $\bmtheta' = [\vecx^T, \bmgm^T]^T$ and hence, $\bmtheta = [{\bmtheta'}^T, \xi]^T$.  In order to compute the HCRB, we need to
find $\mathbf{H}_{\xi}^{\bmtheta}(\xi)$, $\mathbf{H}_{\xi}^{\bmtheta}(\bmtheta')$ and $\mathbf{H}_{\xi}^{\bmtheta}(\bmtheta', \xi)$. We have $\log \pr_{\vecY, \vecX; \bmgm,
\xi}(\vecy, \vecx ; \bmgm, \xi) = \log\pr_{\vecY|\vecX;\xi}(\vecy|\vecx;\xi) + \log \pr_{\vecX; \bmgm}(\vecx ; \bmgm)$, where $\xi = \sigma^{2}$. Using \eqref{Infomat_def_gen}, the
submatrix $\mathbf{H}_{\xi}^{\bmtheta}(\bmtheta') = \mathbf{H}^{\bmtheta'}$, i.e., the same as computed earlier in \eqref{case1_HCRB} when $\bmgm$ is unknown deterministic and by \eqref{bcrb_gamma} when $\bmgm$ is random. Hence, we focus on the block matrices that occur due to the additional parameter $\xi$. First, $\mathbf{H}_{\xi}^{\bmtheta}(\xi)$ is computed as in Sec.~3.6 in \cite{Kay}, from which, $-\mathbb{E}_{\vecY,\vecX;\xi}
\left[-\frac{N}{2\xi^2}\right] = \frac{N}{2\xi^2}$.

From Lemma~\ref{prop_gen_noise},  it directly follows that $\mathbf{H}_{\xi}^{\bmtheta}(\bmgm,\xi) = \mathbf{0}_{L \times 1}$. Using \eqref{Infomat_def_gen}, we compute
$\mathbf{H}_{\xi}^{\bmtheta}(\vecx,\xi)$ as follows:
\begin{eqnarray}
\mathbf{H}_{\xi}^{\bmtheta}(\vecx,\xi) = \mathbb{E}_{\vecX}(\mathbb{E}_{\vecY|\vecX;\xi}(\bmphi^T\vecy - \bmphi^T\bmphi\vecx)).
\label{cross_terms_unk}
\end{eqnarray}
Since $\mathbb{E}_{\vecY|\vecX;\xi}(\vecy) = \bmphi \vecx$, $\mathbb{E}_{\vecX}(\bmphi^T(\bmphi\vecx) - \bmphi^T\bmphi\vecx) = \mathbf{0}_{L \times 1}$. This completes the proof.
\subsection{Proof of Proposition \ref{prop4}}
\label{pf_prop4}
In this case, we define $\bmtheta \triangleq [{\bmtheta'}^T, \xi]$ and $\bmtheta' \triangleq [\vecx^T, \bmgm^T]^T$.  In order to compute the HCRB, we need to find
$\mathbf{H}_{\xi}^{\bmtheta}(\xi)$, $\mathbf{H}_{\xi}^{\bmtheta}(\bmtheta')$ and $\mathbf{H}_{\xi}^{\bmtheta}(\bmtheta', \xi)$. Using \eqref{Infomat_def_gen},  the
expression for $\mathbf{H}_{\xi}^{\bmtheta}(\bmtheta')$ is the same as computed earlier in \eqref{case1_HCRB} when $\bmgm$ is unknown deterministic and by \eqref{bcrb_gamma} when
$\bmgm$ is random. Since $\xi$ is random, the expectation has to be taken over the distribution of $\xi$ also, and hence,
\begin{multline}
\mathbf{H}_{\xi}^{\bmtheta}(\xi) = -\mathbb{E}_{\vecY,\vecX, \Xi}\left[\frac{\partial^2}{\partial \xi^2}(\log \pr_{\vecY|\vecX,\Xi}(\vecy|\vecx,\xi)\right.\\
\left. +\log \pr_\Xi(\xi))\right] 
= \mathbb{E}_{\Xi}\left(\frac{N/2-c-1}{\xi^2}+\frac{2d}{\xi^3}\right).
\end{multline}
The above expectation is evaluated as 
\begin{eqnarray}
&\mathbf{H}_{\xi}^{\bmtheta}(\xi) =  \frac{\left({N}/{2}-c-1\right)d^{c}}{\Gamma(c)} \displaystyle\int\limits_{\xi =
0}^{\infty}\xi^{-2}\xi^{(-c-1)}\exp\left\{-\frac{d}{\xi}\right\}\mathrm{d}\xi + \nonumber\\
&\!\!\!\!\!\!\frac{2d^{(c+1)}}{\Gamma(c)} \int\limits_{\xi = 0}^{\infty}\xi^{-3}\xi^{(-c-1)}\exp\left\{-\frac{d}{\xi}\right\}\mathrm{d}\xi =\frac{c(c+1)\left(\frac{N}{2}+c+3\right)}{d^2}.
\end{eqnarray}
To find the other components of the matrix,  we compute $\mathbf{H}_{\xi}^{\bmtheta}(\bmtheta',\xi) = (\mathbf{H}_{\xi}^{\bmtheta}(\xi, \bmtheta'))^T$, which consists of
$\mathbf{H}_{\xi}^{\bmtheta}(\bmgm,\xi)$ and $\mathbf{H}_{\xi}^{\bmtheta}(\vecx,\xi)$. From Lemma~\ref{prop_gen_noise}, $\mathbf{H}_{\xi}^{\bmtheta}(\bmgm,\xi) = \mathbf{0}_{L \times 1}$. Using the
definition of $\mathbf{H}_{\xi}^{\bmtheta}(\vecx,\xi)$, from \eqref{cross_terms_unk} and since $\pr_\Xi(\xi)$ is not a function of $x_i$, we see that $\mathbf{H}_{\xi}^{\bmtheta}(\vecx,\xi) = (\mathbf{H}_{\xi}^{\bmtheta}(\xi,\vecx))^T = \mathbf{0}_{L \times 1}$. Thus, we obtain the FIM given by \eqref{bcrb_unknoi}.
\subsection{Proof of Theorem \ref{thm4}}
\label{pf_thm3}
First, we show that the log likelihood $\log(\pr_{\vecY;\bmgm, \xi}(\vecy; \bmgm, \xi))$ in \eqref{margi_cost}
satisfies the regularity condition w.r.t. $\xi$. Differentiating the log likelihood w.r.t. $\xi$ and taking $-\mathbb{E}_{\vecY;\bmgm,\xi}(\cdot)$ on both the sides of the
equation,
\begin{eqnarray}
&\frac{\partial}{\partial \xi}\log(\pr_{\vecY;\bmgm, \xi}(\vecy, \bmgm, \xi))  = \frac{1}{2}\frac{\partial}{\partial \xi}(-\log |\bmSigma_y| -
\vecy^T\bmSigma_y^{-1}\vecy)\nonumber\\
& = -\frac{1}{2}\left[\mbox{Tr}(\bmSigma_y^{-1}) -\mbox{Tr}(\vecy\vecy^T(\bmSigma_y^{-1}\bmSigma_y^{-1}))\right], \label{first_diff_lemma4}
\end{eqnarray}
\begin{eqnarray}
&\mathbb{E}_{\vecY;\bmgm,\xi}\left[\mbox{Tr}(-\frac{1}{2}\bmSigma_y^{-1}) +\frac{1}{2}\mbox{Tr}(\vecy\vecy^T(\bmSigma_y^{-1}\bmSigma_y^{-1}))\right]  \nonumber\\
&= \frac{1}{2}\left[\mbox{Tr}(\bmSigma_y^{-1}) - \mbox{Tr}(\bmSigma_y^{-1})\right] = 0.
\end{eqnarray}
Hence, the regularity condition is satisfied. From \eqref{MCRB_detgm},  we have $(\mathbf{M}_{\xi}^{\bmtheta}(\bmgm))_{ij} = - \frac{(\Phi_j^T \bmSigma_y^{-1}\Phi_i)^2}{2}$. To obtain
$\mathbf{M}_{\xi}^{\bmtheta}(\xi)$, we differentiate \eqref{first_diff_lemma4} w.r.t. $\xi$ to obtain
\begin{eqnarray}
\frac{\partial^2}{\partial \xi^2}(\log \pr_{\vecY;\bmgm,\xi}(\vecy;\bmgm,\xi)) = \frac{1}{2}\mbox{Tr}(\bmSigma_y^{-2})- \mbox{Tr}(\vecy\vecy^T(\bmSigma_y^{-3})).
\end{eqnarray}
Taking $-\mathbb{E}_{\vecY;\bmgm,\xi}(\cdot)$ on both sides of the above equation, 
\begin{eqnarray}
&\mathbf{M}_{\xi}^{\bmtheta}(\xi) = -\mathbb{E}_{\vecY;\bmgm,\xi}\left[\frac{1}{2}\mbox{Tr}(\bmSigma_y^{-2})- \mbox{Tr}(\vecy\vecy^T\mbox{Tr}(\bmSigma_y^{-3}))\right]   \nonumber\\
&= \mbox{Tr}(\bmSigma_y^{-2}) - \frac{1}{2}\mbox{Tr}(\bmSigma_y^{-2}) = \frac{1}{2}\mbox{Tr}(\bmSigma_y^{-2}).
\label{mcrb_xi}
\end{eqnarray}
The vector $\mathbf{M}_{\xi}^{\bmtheta}(\bmgm,\xi)$ is found by differentiating \eqref{firstdiff_margam} w.r.t. $\xi$ and taking the negative expectation:
\begin{multline}
(\mathbf{M}_{\xi}^{\bmtheta}(\bmgm,\xi))_i   \\
= \mathbb{E}_{\vecY;\bmgm,\xi}\left[\frac{\partial}{\partial\xi}\left(\frac{\Phi_i^T \bmSigma_y^{-1} \Phi_i -
\Phi_i^T\bmSigma_y^{-1}\vecy\vecy^T\bmSigma_y^{-1}\Phi_i}{2}\right)\right] \\ 
 = \frac{1}{2}\Phi_i^T\bmSigma_y^{-2}\Phi_i.
\label{crossterm_1}
\end{multline}
Since $\mathbf{M}_{\xi}^{\bmtheta}(\xi, \bmgm)  = (\mathbf{M}_{\xi}^{\bmtheta}(\bmgm,\xi))^T$, the $i^{\text{th}}$ term of $(\mathbf{M}_{\xi}^{\bmtheta}(\xi, \bmgm))_i =
\frac{1}{2}\Phi_i^T\bmSigma_y^{-2}\Phi_i$. The MCRB $\mathbf{M_{\xi}^{\bmtheta}}$ can now be obtained by  combining the expressions in
\eqref{MCRB_detgm}, \eqref{mcrb_xi} and \eqref{crossterm_1}; this completes the proof.

\bibliographystyle{IEEEtran}
\bibliography{mybib_hcrb}
\begin{biography}[{\includegraphics[width=1in,height=2.25in,clip,keepaspectratio]{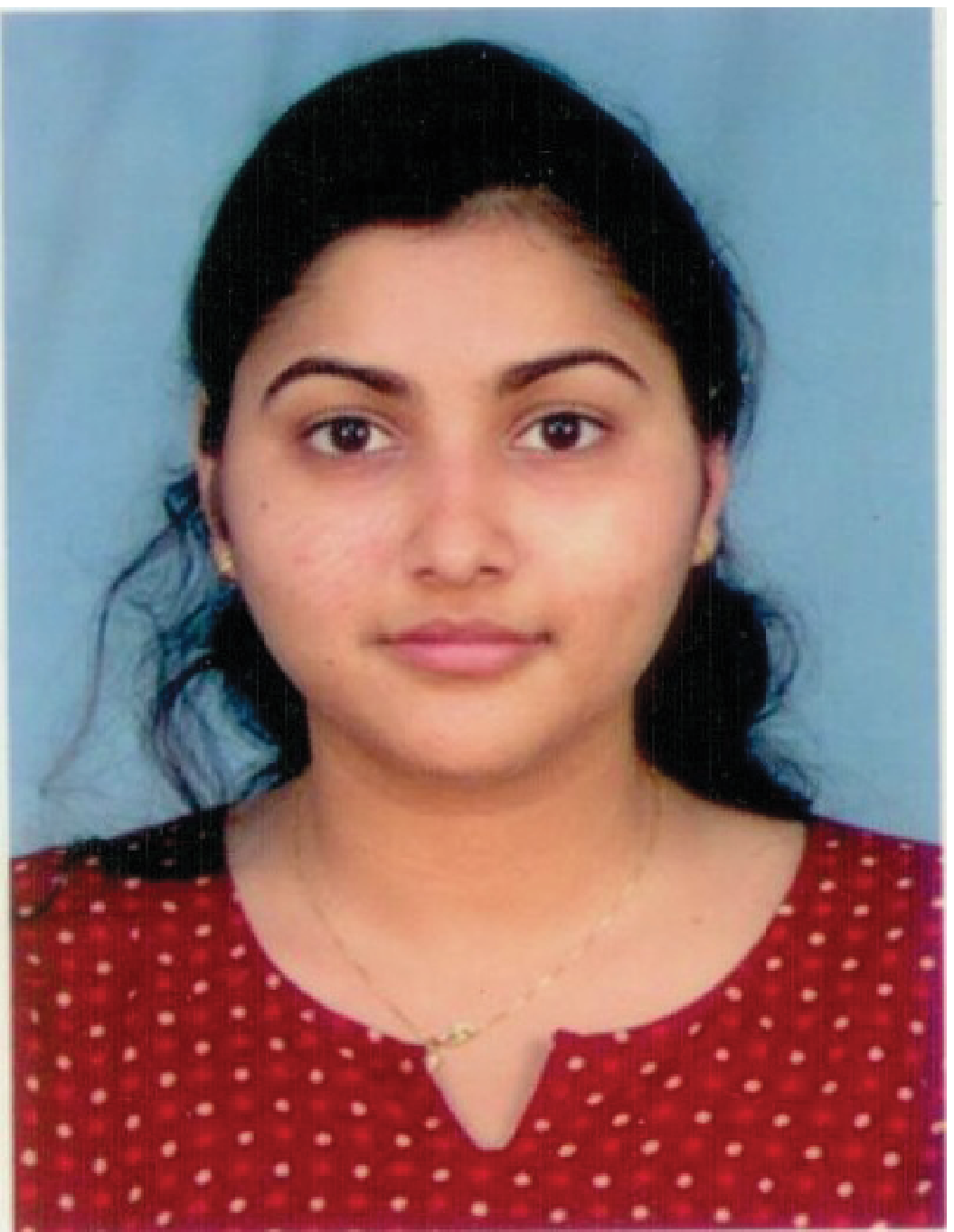}}]{Ranjitha Prasad} received the B.E. degree in Electronics and Communication Engineering from National Institute of Engineering, Mysore, India, in 2004, and the M.S. degree in Electrical Engineering, Indian Institute of Technology Madras, Chennai, India, in 2009. From July 2004-2006, she worked as a senior design engineer at Tata Elxsi, Bangalore, India. She is currently working towards the Ph.D degree at the Department of Electrical Communication Engineering, Indian Institute of Science, Bangalore, India. Her research interests include signal processing for communications, adaptive filter theory, sparse Bayesian learning and compressive Sensing.
\end{biography}

\begin{biography}[{\includegraphics[width=1in,height=2.25in,clip,keepaspectratio]{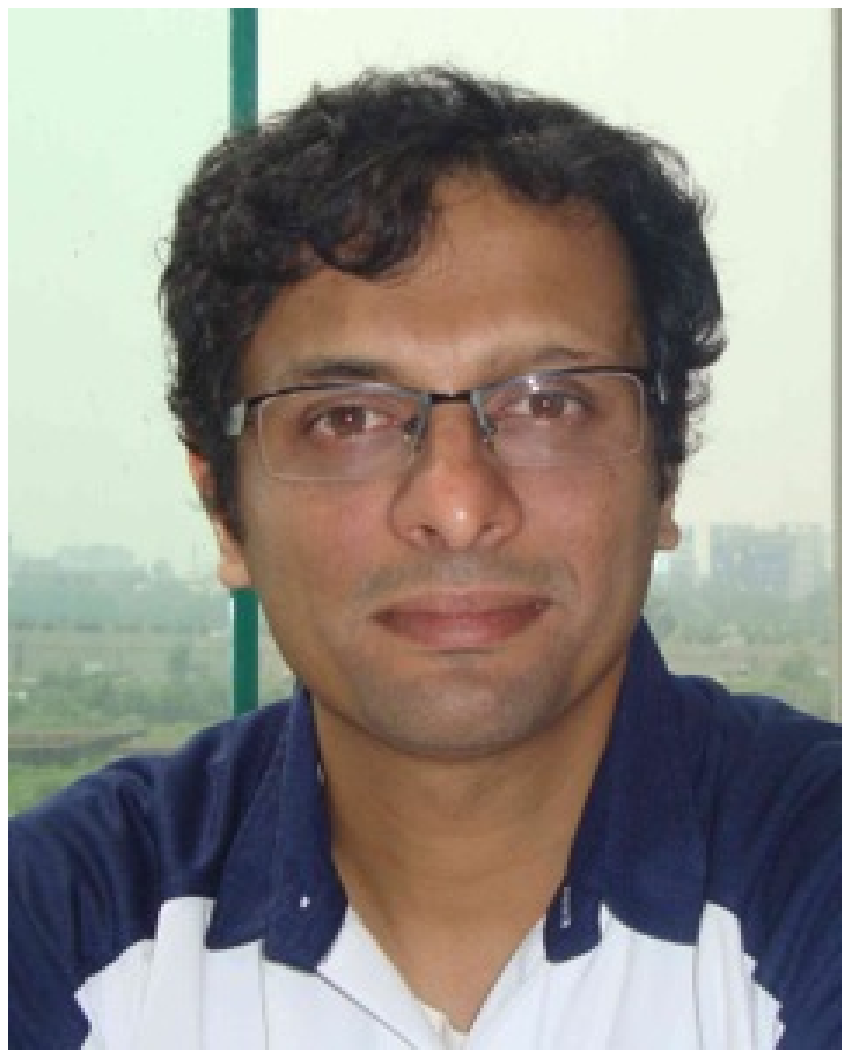}}]{Chandra R. Murthy} (S'03--M'06 -- SM'11) received the B.Tech. degree in Electrical Engineering from the Indian Institute of Technology Madras, Chennai, India, in 1998, the M.S. and Ph.D. degrees in Electrical and Computer Engineering from Purdue University, West Lafayette, IN and the University of California, San Diego, CA, in 2000 and 2006, respectively. 

From 2000 to 2002, he worked as an engineer for Qualcomm Inc., San Jose, CA, where he worked on WCDMA baseband transceiver design and 802.11b baseband receivers. From Aug. 2006 to Aug. 2007, he worked as a staff engineer at Beceem Communications Inc., Bangalore, India on advanced receiver architectures for the 802.16e Mobile WiMAX standard. In Sept. 2007, he joined as an assistant professor at the Department of Electrical Communication Engineering at the Indian Institute of Science, Bangalore, India, where he is currently working. His research interests are in the areas of Cognitive Radio, Energy Harvesting Wireless Sensors and MIMO systems with channel-state feedback. He is currently serving as an associate editor for the IEEE Signal Processing Letters.
\end{biography}

\end{document}